\documentclass[sigconf,nonacm]{acmart}

\AtBeginDocument{%
  }

\usepackage{graphicx}
\graphicspath{{./}{figures/}}
\usepackage{pifont}
\usepackage{xcolor}
\usepackage{colortbl}
\usepackage{booktabs}
\usepackage{multirow}
\usepackage{amsmath}
\usepackage{balance}
\usepackage{dblfloatfix}
\usepackage{placeins}
\usepackage{tcolorbox}
\tcbuselibrary{skins,breakable}

\definecolor{regroundpurple}{RGB}{232, 222, 248}
\definecolor{pipelinepurple}{RGB}{213, 200, 232}
\definecolor{correctionred}{RGB}{244, 191, 191}
\definecolor{groundingyellow}{RGB}{244, 224, 175}
\definecolor{verificationgreen}{RGB}{182, 226, 200}
\newtcolorbox{promptbox}[2][]{%
  colback=#2!12, colframe=#2!55!black,
  fonttitle=\bfseries\sffamily\footnotesize,
  title=#1,
  boxrule=0.4pt, arc=2pt,
  left=5pt, right=5pt, top=4pt, bottom=4pt,
  fontupper=\footnotesize
}
\begin{document}

\title{ReGround: Restoring Visual Grounding in Multi-Step Reasoning through Self-Diagnosis and Visual Re-Examination}

\author{Lei Peng}
\authornote{These authors contributed equally.}
\orcid{0009-0006-4334-899X}
\affiliation{%
  \institution{University of Science and Technology of China}
  \department{School of Artificial Intelligence and Data Science}
  \city{Hefei}
  \state{Anhui}
  \country{China}}
\email{peng.lei@mail.ustc.edu.cn}

\author{Shuai Lv}
\authornotemark[1]
\authornote{Corresponding authors.}
\orcid{0000-0002-1154-3612}
\affiliation{%
  \institution{University of Science and Technology of China}
  \department{School of Artificial Intelligence and Data Science}
  \city{Hefei}
  \state{Anhui}
  \country{China}}
\email{shuailv@mail.ustc.edu.cn}

\author{Wei Hu}
\authornotemark[2]
\orcid{0000-0001-9629-2121}
\affiliation{%
  \institution{University of Science and Technology of China}
  \department{School of Artificial Intelligence and Data Science}
  \city{Hefei}
  \state{Anhui}
  \country{China}}
\affiliation{%
  \institution{University of Science and Technology of China}
  \department{State Key Laboratory of Precision and Intelligent Chemistry}
  \city{Hefei}
  \state{Anhui}
  \country{China}}
\email{whuustc@ustc.edu.cn}

\renewcommand{\shortauthors}{Peng et al.}

\begin{abstract}
Vision-Language Models (VLMs) often lose visual grounding during multi-step reasoning: as reasoning chains grow longer, later inference steps rely increasingly on language priors rather than image evidence.
We identify a consistent benchmark-level signature associated with this degradation: across 2{,}510 re-examined samples from four benchmarks, attention entropy over image tokens typically decreases during Round~1 and rises again after image re-injection.

However, we find that effective visual re-examination requires two complementary ingredients: image re-injection and targeted self-diagnosis.
Without targeted diagnosis, re-examination can even hurt performance, whereas accurate self-diagnosis yields substantial gains---a swing of several points on key benchmarks, indicating that diagnostic quality is a key factor in whether re-examination helps or hurts in our setting.
We present \textbf{ReGround}, a two-stage framework that teaches VLMs to self-diagnose grounding failures and selectively re-examine visual evidence, without architectural modifications or external tools.
Through \emph{capability bootstrapping}, a stronger variant from the same model family provides diagnostic scaffolding only during data construction, while the policy model learns to diagnose autonomously at inference time and retains most of the assisted gains.

Experiments on eight benchmarks across two VLM backbones demonstrate consistent gains, especially on visually intensive multi-step reasoning tasks, while incurring only modest inference overhead relative to tool-augmented baselines.
Project page: \url{https://sespoir.github.io/reground-page/}. Code: \url{https://github.com/sespoir/ReGround}.
\end{abstract}



\maketitle

\section{Introduction}
\label{sec:intro}

Vision-Language Models (VLMs) have advanced multimodal reasoning~\cite{qwen25vl, qwen3vl, gpt4o} and increasingly power interactive agents that carry out long-horizon, vision-grounded tasks~\cite{zhang2026omegause}, yet their reliance on visual evidence can weaken during long reasoning chains as language priors increasingly dominate. We call this \emph{visual grounding decay}: attention to image tokens progressively narrows as textual context accumulates. Prior work reports this pattern through depth-wise attention decay~\cite{fastv, kaduri2024}, declining image--text mutual information~\cite{m3id}, and visual attention sinks~\cite{visual-attention-sink}. The problem is especially acute in mathematical and scientific reasoning, where every step must remain anchored to image-specific values and spatial relations.

\begin{figure*}[t]
  \centering
  \includegraphics[width=\textwidth]{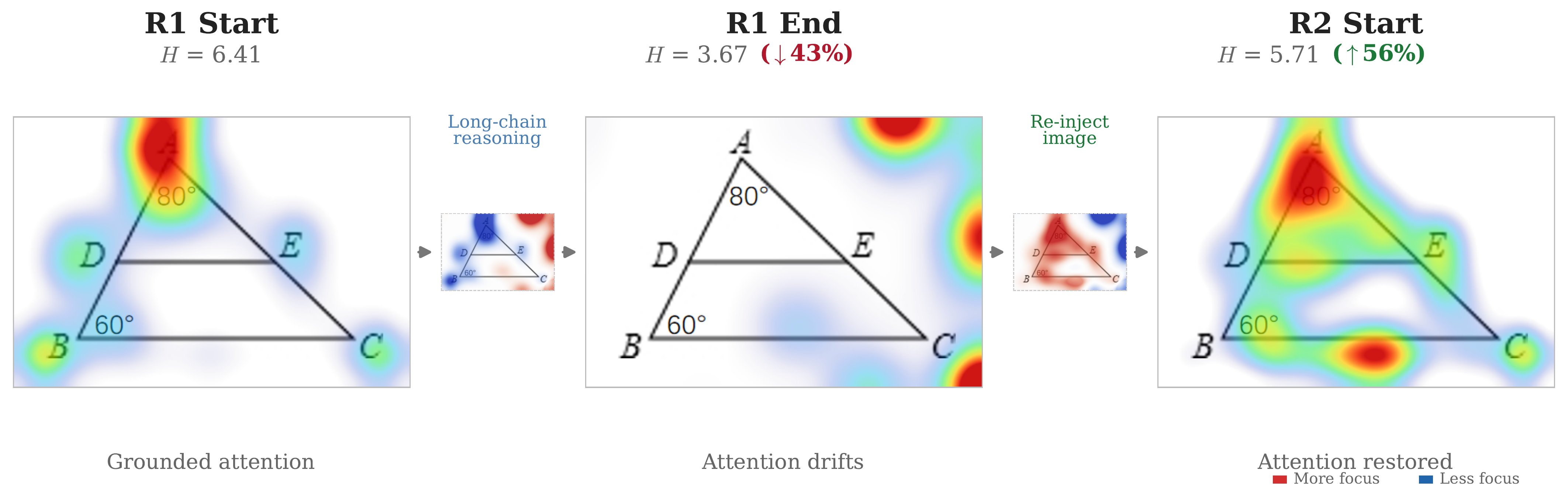}
  \caption{
    \textbf{Attention distribution narrows during multi-step reasoning and is redistributed after image re-injection.}
    Qwen2.5-VL-7B attention on a MathVerse example narrows from \emph{R1 Start} ($H{=}6.41$) to \emph{R1 End} ($H{=}3.67$, $\downarrow$43\%), then redistributes after the same image tokens are re-presented in Round~2 ($H{=}5.71$, $\uparrow$56\%). Spatial entropy measures attention breadth and is a proxy rather than a direct grounding metric. Re-examination changes the initially incorrect prediction to the correct answer; insets show increases (red) and decreases (blue) between rounds.
  }
  \label{fig:motivation}
  \Description{Three attention heatmaps on a geometry diagram showing attention narrowing during Round 1 reasoning and redistribution after image re-injection in Round 2.}
\end{figure*}

Figure~\ref{fig:motivation} illustrates the effect: spatial entropy over image tokens drops from $H{=}6.41$ to $H{=}3.67$ during Round~1, then rises to $H{=}5.71$ after image re-injection. Across 2{,}510 samples from four benchmarks, entropy decreases during Round~1 in 98.0\% of cases and rebounds after re-injection in 94.4\% (\S\ref{sec:visual_evidence}). A separate behavioral ablation shows much smaller gains from text-only reflection than from visual re-injection (Table~\ref{tab:core_ablation}, rows \ding{174} vs.\ \ding{175}). We therefore use entropy as a mechanism-level signature, not a direct grounding metric.

Reversibility alone does not guarantee correction. Re-injection re-presents the same visual tokens at new sequence positions after the model's reasoning and diagnostic cue, allowing updated context to redirect cross-modal attention. The result is a targeted recalibration rather than a full reset ($H_{\mathrm{R2S}} < H_{\mathrm{R1S}}$). Yet indiscriminate re-examination can trigger harmful second-guessing (\S\ref{sec:data_ablation}); the model must learn both \emph{when} to re-examine and \emph{what} to revisit.

Diagnostic quality---\emph{what} to re-examine---is the main bottleneck in our setting. A generic cue can underperform no re-examination, whereas targeted self-diagnosis yields substantial gains (Table~\ref{tab:cue_ablation}). Without guidance, a second round may override a correct answer; with diagnosis, updated context emphasizes neglected evidence. Thus effective re-examination requires both image re-injection and a meaningful cue, distinguishing ReGround from methods centered on correction mechanisms alone~\cite{thyme, deepeyes, qwen-lookagain, look-back}.

We present \textbf{ReGround}, a two-stage, tool-free framework. Stage~1 routes incorrect, visually ungrounded, or stochastically selected samples to visual self-diagnosis and image re-injection (Figure~\ref{fig:pipeline}). Through \emph{capability bootstrapping}, a stronger model variant supplies diagnostic cues only during data construction; the policy learns to generate them autonomously. Stage~2 uses GRPO~\cite{deepseek-r1} with an asymmetric reward to calibrate triggering. ReGround targets grounding drift in visually intensive multi-step reasoning without architectural changes.

Our core contributions are summarized as follows:
\begin{itemize}
    \item \textbf{Empirical evidence of visual grounding decay and recovery.}
    Across 2{,}510 samples, attention entropy narrows during reasoning in 98.0\% of cases and usually recovers after re-injection.

    \item \textbf{Diagnostic quality as the key bottleneck.}
    Generic re-examination can hurt, while self-diagnosis recovers 79--90\% of the oracle cue's gains without inference-time dependency.
    \item \textbf{Triggering policy analysis.}
    Over-triggering damages correct answers, whereas conservative triggering forgoes recoverable gains (\S\ref{sec:data_ablation}).

    \item \textbf{A practical, tool-free framework.}
    ReGround improves eight benchmarks and two VLM backbones without external tools or architectural changes.
\end{itemize}


\section{Related Work}
\label{sec:related}

\subsection{Reasoning Training for VLMs}
\label{sec:rw_reasoning}

Recent efforts to enhance VLM reasoning follow two main lines.
The first distills structured reasoning traces from stronger models.
LLaVA-CoT~\cite{llava-cot} trains multistage reasoning---summarization, visual interpretation, logical reasoning, and conclusion---on 100K structured annotations.
Chen et al.~\cite{chen2025sftorl} study SFT-versus-RL training for reasoning LVLMs and introduce the VLAA-Thinker family, while OpenVLThinker~\cite{openvlthinker} develops iterative SFT--RL cycles for complex visual reasoning.
While effective, these approaches rely on teacher-generated reasoning traces that may differ from the student's native rollout distribution, encouraging imitative rather than self-discovered reasoning~\cite{chen2025sftorl,openvlthinker}.

The second line applies reinforcement learning directly.
Following DeepSeek-R1~\cite{deepseek-r1}, which showed that GRPO alone can induce chain-of-thought behavior, MM-Eureka~\cite{mm-eureka} and VLM-R1~\cite{vlm-r1} extend rule-based RL to multimodal reasoning, while VL-Rethinker~\cite{vl-rethinker} identifies the ``vanishing advantages'' problem in GRPO and proposes selective sample replay to sustain exploration.

ReGround shares the two-stage SFT$\to$RL paradigm but differs in emphasis: rather than distilling \emph{answer generation} ability, we distill \emph{visual diagnostic} ability, while keeping all reasoning tokens on the policy model's own generation distribution.

\subsection{Visual Grounding Decay in VLMs}
\label{sec:rw_diminishing}

A growing body of work documents that VLMs progressively lose contact with visual evidence during extended generation.
FastV~\cite{fastv} observes that attention to image tokens decays rapidly after the first few decoding steps and proposes pruning low-attention visual tokens to accelerate inference.
M3ID~\cite{m3id} formalizes this from a mutual-information perspective, showing that the conditional mutual information between generated tokens and the image decreases with generation length.
Kaduri et al.~\cite{kaduri2024} find that attention to image tokens is concentrated in early layers and diminishes in deeper ones, with query tokens acting as compressed global image descriptors, and See What You Are Told~\cite{visual-attention-sink} further identifies visual attention-sink behavior in which textual context increasingly dominates attention allocation.

These findings collectively motivate approaches that re-expose visual evidence during extended reasoning. Our work builds on this line but goes further: rather than only documenting the decay, we ask what determines whether visual revisitation is actually effective, and find that diagnostic quality plays a key role.

\subsection{Visual Self-Correction in VLMs}
\label{sec:rw_multiturn}

Self-correction in language models has been extensively studied, with broad consensus that LLMs struggle to improve their own outputs without external feedback~\cite{self-refine, intrinsic-selfcorrect}. In the multimodal setting, recent work explores two main strategies.

\noindent\textbf{Tool-augmented approaches.}
DeepEyes~\cite{deepeyes} learns to invoke image-cropping tools via end-to-end RL, though DeepEyesV2~\cite{deepeyesv2} reports that pure RL suffers from tool-calling collapse, ultimately requiring cold-start SFT.
Thyme~\cite{thyme} enables VLMs to generate and execute Python code for image manipulation and mathematical computation, requiring 500K SFT samples and a custom GRPO-ATS algorithm.
While achieving strong results on fine-grained perception, these methods require tool environments at deployment and introduce format constraints that complicate cross-task generalization.

\noindent\textbf{Implicit re-focusing.}
Look-Back~\cite{look-back} introduces a \texttt{<back>} token that encourages the model to re-focus attention on visual inputs without explicitly re-injecting images. While elegant, this approach re-attends within the late, text-dominated context where attention to visual tokens has already narrowed.

\noindent\textbf{Our distinction.}
ReGround differs from tool-augmented methods in requiring no external tools---only re-injection of the same image in a new dialogue turn. It differs from implicit re-focusing by re-presenting the visual tokens at new sequence positions after the diagnostic context rather than re-attending to them only at the end of the original reasoning chain. Most importantly, while Look-Back~\cite{look-back}, Qwen-LookAgain~\cite{qwen-lookagain}, and Reflection-V~\cite{reflection-v} also employ multi-stage training with visual re-examination, none isolates whether performance gains stem from the \emph{format} of visual re-examination or from the \emph{quality} of diagnostic guidance within it. Our cue-quality ablation (Table~\ref{tab:cue_ablation}) provides this missing comparison, revealing that diagnostic quality---not the re-examination format alone---is the key variable. A complementary mechanism ablation (Table~\ref{tab:core_ablation}) further confirms that high-quality training data without visual re-injection yields negligible gains, establishing that both ingredients are necessary.


\begin{figure*}[t]
\centering
\includegraphics[width=\textwidth]{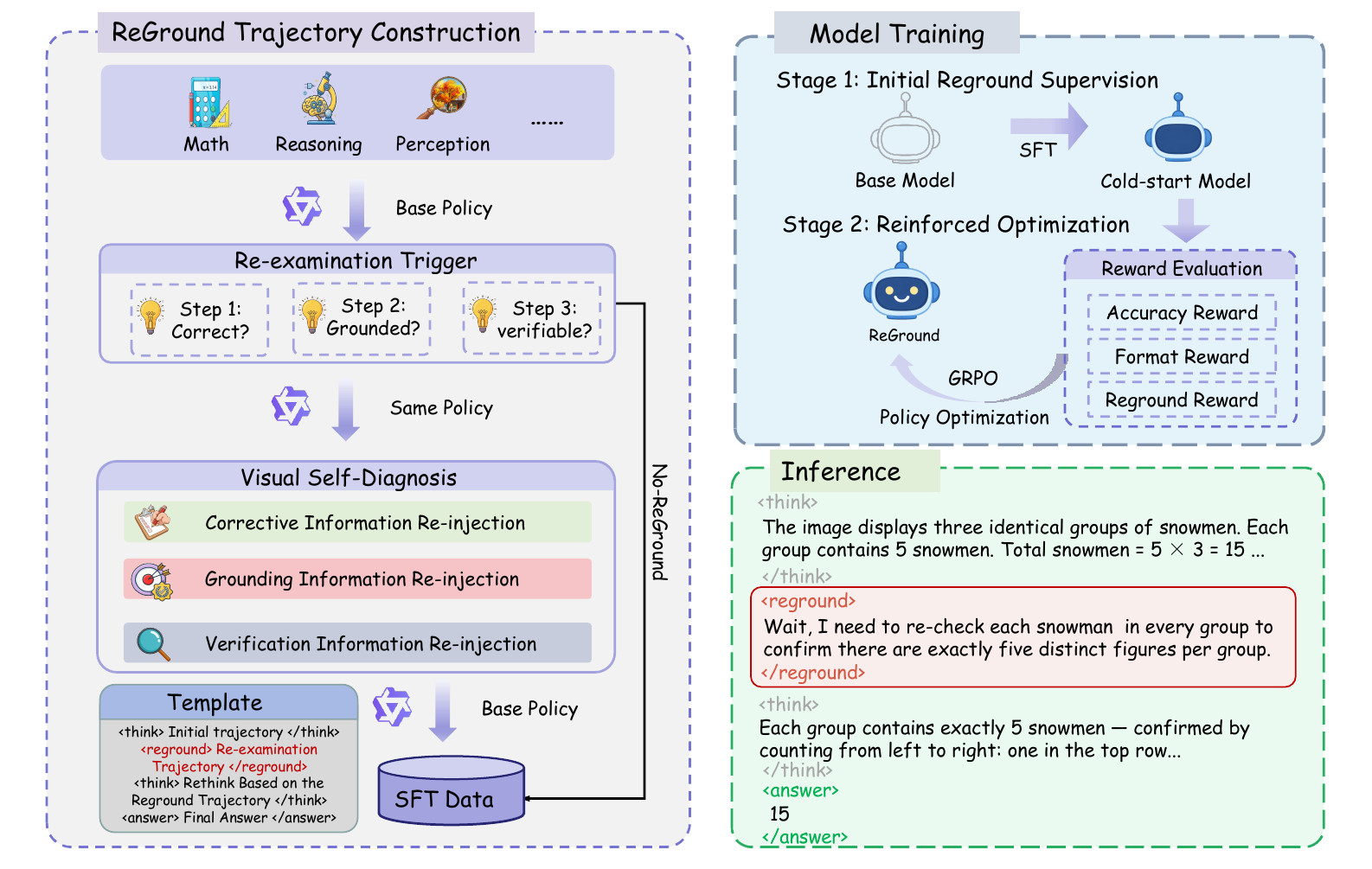}
\caption{Overview of the ReGround framework.
\textbf{Trajectory Construction (left):} Initial reasoning is evaluated by a sequential Re-examination Trigger. Triggered samples undergo Visual Self-Diagnosis for information re-injection, forming formatted SFT data.
\textbf{Model Training (upper right):} Stage~1 applies SFT for cold-start initialization. Stage~2 uses GRPO with a multi-objective reward to refine the triggering policy.
\textbf{Inference (lower right):} The model autonomously generates a \texttt{<reground>} cue to trigger visual re-injection before producing the revised reasoning and final \texttt{<answer>}.}
\Description{Pipeline showing trajectory construction via triggering and self-diagnosis (left), two-stage training with SFT and GRPO (upper right), and autonomous inference (lower right).}
\label{fig:pipeline}
\end{figure*}

\section{Method}
\label{sec:method}

We present \textbf{ReGround}, a two-stage framework that teaches VLMs to autonomously re-examine visual evidence and revise their reasoning.
Stage~1 bootstraps trajectories and trains the model via SFT (\S\ref{sec:sft}).
Stage~2 refines the triggering policy through GRPO (\S\ref{sec:grpo}).
Figure~\ref{fig:pipeline} provides an overview.

\subsection{ReGround Trajectory Construction for Cold-Start SFT}
\label{sec:sft}
As illustrated in Figure~\ref{fig:pipeline}, the construction pipeline operates on diverse multimodal datasets and comprises two core modules: a sequential Re-examination Trigger and a Visual Self-Diagnosis mechanism.
\begin{figure*}[t]
\centering
\includegraphics[width=\textwidth]{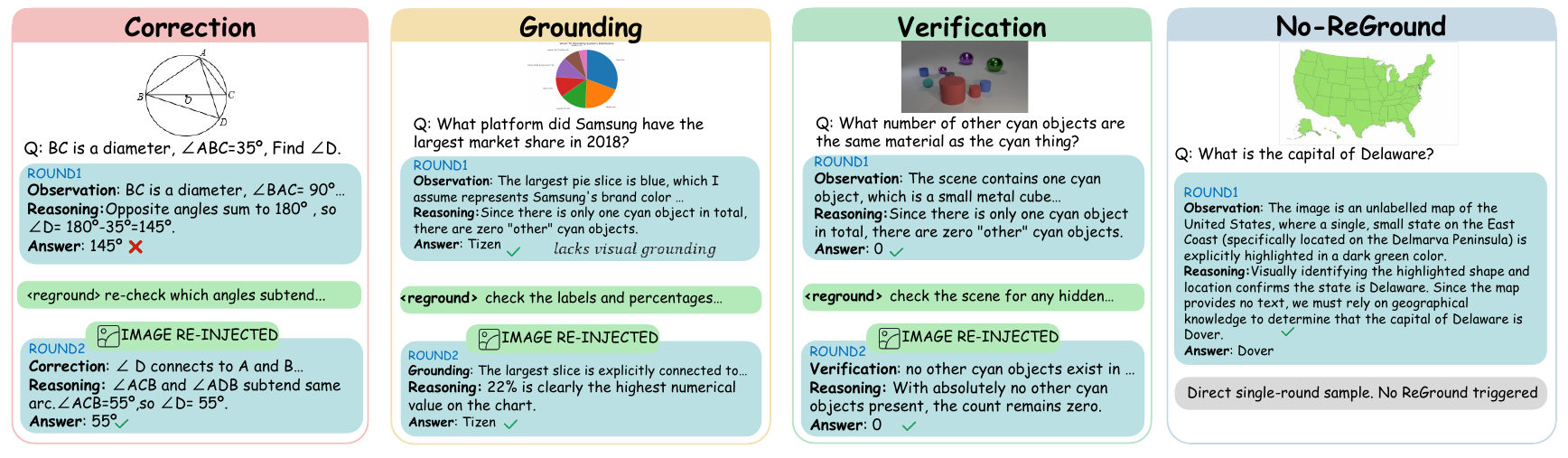}
\caption{Representative trajectories for each trajectory mode.
\textbf{Correction (left):} Round~1 misidentifies the arc relationship (145\textdegree). After ReGround, Round~2 correctly derives 55\textdegree.
\textbf{Grounding (center-left):} Round~1 answers ``Tizen'' correctly but relies on color association rather than explicit labels. Round~2 grounds the answer in ``22\%.''
\textbf{Verification (center-right):} Round~1 correctly answers 0. The $\varepsilon$-trigger initiates re-examination, and Round~2 systematically confirms.
\textbf{No-ReGround (right):} ``Dover'' is answered correctly and with visual grounding in a single round; no re-examination is triggered.}
\Description{Four trajectories in a row showing Correction, Grounding, Verification, and No-ReGround modes.}
\label{fig:trajectories}
\end{figure*}
\subsubsection{Re-examination Trigger Policy.}
\label{sec:trigger}

Not every sample benefits equally from re-examination. To prevent the model from learning a degenerate always-trigger behavior, we design a stepwise Re-examination Trigger that filters the Base Policy's initial response through three sequential conditions:

\noindent\textbf{Step 1: Is Correct?}
We first evaluate the accuracy of the initial response. If the answer is incorrect, the trajectory unconditionally triggers re-examination to encourage recovery from actual visual failure.

\noindent\textbf{Step 2: Is Grounded?}
If the answer is correct, it may still emerge from language priors rather than genuine image understanding~\cite{qwen-lookagain}. We assess the initial reasoning against four operational grounding criteria:

\begin{enumerate}
\item~\emph{Visual attribute coverage:} explicit identification of task-relevant visual marks;
\item~\emph{Alternative elimination:} ruling out competing options via visual evidence;
\item~\emph{Hedging absence:} avoidance of speculative language;
\item~\emph{Visual anchor:} tying the final answer to concrete visual elements.
\end{enumerate}

Failure on any criterion triggers re-examination. We implement this routing decision with Qwen2.5-VL-72B-Instruct, which receives the question, image, and full Round~1 response and returns structured pass/fail decisions for the four criteria. Among correct Round~1 responses, 22.9\% are classified as visually ungrounded. This judge is used for Stage~1 data construction and is not required during GRPO, inference, or benchmark scoring; evaluating its agreement with human annotations remains future work.

\noindent\textbf{Step 3: Is Verifiable?}
For well-grounded, correct answers, we apply a stochastic trigger with probability $\varepsilon$. This injects verification behavior, demonstrating that re-examination can serve to confirm correct reasoning, preventing the misconception that re-examination is only warranted when something is wrong.

\noindent\textbf{No-ReGround.}
Samples passing Step 1 and 2 without hitting the Step 3 stochastic trigger bypass re-examination entirely. They are retained as single-turn trajectories, providing the negative examples necessary for a balanced triggering policy, fostering restraint.

\subsubsection{Visual Self-Diagnosis and Re-injection.}
\label{sec:reinjection}

Once re-examination is triggered, the system initiates the Visual Self-Diagnosis module.
Figure~\ref{fig:trajectories} illustrates the resulting Correction, Grounding, Verification, and No-ReGround trajectory modes:

\begin{itemize}
    \item \textbf{Corrective Information Re-injection (Step~1):} Focuses on targeted error recovery. The injected cue identifies the most likely point of visual misperception so that the model can redirect attention to the relevant visual tokens.
    \item \textbf{Grounding Information Re-injection (Step~2):} Focuses on visual alignment. The model is forced to anchor its speculative language to concrete spatial coordinates or visual elements, mitigating language hallucination. The cue specifies the relation to inspect without revealing the target answer, so the model must still resolve the evidence itself.
    \item \textbf{Verification Information Re-injection (Step~3):} Focuses on consistency. The image is re-injected holistically to double-check the logical chain, enhancing high-confidence robustness. This mode applies when the initial answer appears plausible but its supporting visual chain remains uncertain.
\end{itemize}

\noindent\textbf{Why Visual Re-injection?} We contrast our mechanism with two alternatives.
\emph{(a) Text-only reflection} asks the model to self-correct by re-reading its textual summary. Once critical visual details are lost, text reflection cannot recover them.
\emph{(b) Implicit re-focusing}~\cite{look-back} uses a special token to re-attend to existing visual representations.
With a frozen deterministic visual encoder, identical preprocessing yields the same visual tokens; image re-injection places these tokens at a new sequence position after the diagnostic cue, enabling renewed attention under the updated context.

\subsubsection{Trajectory Formatting and Capability Bootstrapping.}
\label{sec:formatting}

To unify these trajectories for SFT, we formulate a standardized multi-turn template (shown in Figure~\ref{fig:pipeline}, bottom left):

\begin{enumerate}
    \item \texttt{<think>} Initial trajectory \texttt{</think>} (Initial reasoning)
    \item \texttt{<reground>} Re-examination Trajectory \texttt{</reground>} (Visual Self-Diagnosis and visual re-injection)
    \item \texttt{<think>} Rethink Based on the Reground Trajectory \texttt{</think>} (Revised reasoning)
    \item \texttt{<answer>} Final Answer \texttt{</answer>}
\end{enumerate}

During data construction, we employ capability bootstrapping: a larger variant from the same model family generates the diagnostic cues inside the \texttt{<reground>} tag, owing to its stronger instruction-following capability. It provides a first-person reflection (e.g.,
\emph{``Wait, I need to re-examine whether ...''}) without revealing the final answer. After supervised fine-tuning, the policy model learns to produce these cues autonomously, and the larger variant is discarded.
Overall, this multi-turn template encourages the model to revisit the image conditioned on its own intermediate reasoning, producing a structured visual re-examination process driven by self-diagnosis. The retained trajectories form a filtered supervision set $\mathcal{D}_{\text{cold}}$.

\subsection{Model Training}
\label{sec:grpo}
\subsubsection{Cold-start SFT} After generating ReGround trajectories, we retain only those rollouts $\tau \in \mathcal{T}(s)$ whose final prediction satisfies $\tau_{\text{ans}} = a_s$. The training objective is as follows:

\begin{equation}
\mathcal{L}_{\text{cold-start}} = -\mathbb{E}_{(s,\tau)\sim\mathcal{D}_{\text{cold}}} \sum_{t=1}^{|\tau|} \log \pi_\theta(\tau_t \mid i, q, \tau_{<t}),
\end{equation}

This yields the cold-start model $\pi_{\text{SFT}}$, which is endowed with visual re-examination capability.

\subsubsection{GRPO and Reward Evaluation.}
\label{sec:reward}

We optimize the policy on the RL data using Group Relative Policy Optimization (GRPO),
sampling a group of $N$ trajectories $\{\tau_1, ..., \tau_N\}$ per query.
The GRPO objective maximizes the relative advantage while constraining the KL divergence against the Cold-start model.

To comprehensively assess the multi-step reasoning trajectory, the total reward function $\mathcal{R}(\tau_i)$ is formulated as a weighted combination of three components:
\begin{equation}
\label{eq:reward}
\mathcal{R}(\tau_i) = \lambda_{reg} R_{reg}(\tau_i) + \lambda_{acc} R_{acc}(\tau_i) + \lambda_{form} R_{form}(\tau_i)
\end{equation}

The coefficients $\lambda_{reg}$, $\lambda_{acc}$, and $\lambda_{form}$ dictate the relative importance of re-grounding behavior, accuracy, and format adherence (exact hyperparameter values and tuning details are provided in the supplementary material). We use two deterministic binary indicators: $\mathbb{I}_{reg} \in \{0, 1\}$ for a structurally valid re-grounding action and $\mathbb{I}_{acc} \in \{0, 1\}$ for final-answer correctness.

\noindent\textbf{1. Re-grounding Reward ($R_{reg}$):}
We set $R_{reg}(\tau_i)=\mathbb{I}_{reg}$. The indicator is one when the trajectory contains a single non-empty, well-formed \texttt{<reground>} span at the required position, and zero otherwise.

\noindent\textbf{2. Accuracy Reward ($R_{acc}$):}
We compute $\mathbb{I}_{acc}$ by applying the deterministic \texttt{answers\_match} comparison to predictions and targets normalized by the VLMEvalKit parser used for final scoring. To balance the reward distribution, we introduce a negative interaction term between re-examination and accuracy, governed by hyperparameter scales $\gamma$ and $\beta$:
\begin{equation}
R_{acc}(\tau_i) = \mathbb{I}_{acc} - \gamma (\mathbb{I}_{reg} \cdot \mathbb{I}_{acc}) - \beta
\end{equation}
This implies that while correctness is rewarded, its marginal benefit slightly decreases if the model has already obtained the intrinsic reward for initiating a diagnostic re-examination ($R_{reg}$).

\noindent\textbf{3. Formatting Penalty ($R_{form}$):}
Acts as a strict constraint against syntax deviations. $R_{form}(\tau_i) = -1$ if the trajectory exhibits formatting failures (e.g., unbalanced \texttt{<think>}, \texttt{<reground>}, or \texttt{<answer>} tags, or excessive answer length); otherwise, $R_{form}(\tau_i) = 0$.

\begin{table*}[t]
\centering
\small
\caption{Main results on multimodal benchmarks. Best in \textbf{bold}; second-best \underline{underlined}.}
\label{tab:main_results}
\setlength{\tabcolsep}{5pt}
\begin{tabular}{llcccccccc}
\toprule
\textbf{Model} & \textbf{Method} & \textbf{MathVista} & \textbf{MathVision}
& \textbf{MathVerse} & \textbf{MMBench} & \textbf{MMStar}
& \textbf{Hallus.} & \textbf{VisuLogic} & \textbf{V*Bench} \\
\midrule
\multicolumn{10}{l}{\textit{Proprietary Models (reference)}} \\
\midrule
GPT-4o & --- & 63.8 & 30.3 & 39.4 & 83.4 & 63.9 & 69.3 & 26.7 & 72.2 \\
\midrule
\multicolumn{10}{l}{\textit{Open-Source 7--8B Models}} \\
\midrule
Qwen2.5-VL-7B & Base & 68.2 & 27.0 & 35.5 & 82.6 & 64.7 & 64.7 & 20.0 & 78.2 \\
Qwen2.5-VL-7B & R1-VL~\cite{r1-vl} & 65.7 & 24.7 & 37.6 & 83.1$^\dagger$ & 60.0 & 68.8 & 22.3 & 81.2$^\dagger$ \\
Qwen2.5-VL-7B & Thyme~\cite{thyme} & 70.0 & 27.6 & \underline{39.1} & 83.4$^\dagger$ & 65.9 & \underline{71.0} & 23.4 & \textbf{83.5} \\
Qwen2.5-VL-7B & Look-Back~\cite{look-back} & \underline{70.8} & \underline{28.4} & 38.9 & \textbf{83.9}$^\dagger$ & \underline{66.8}$^\dagger$ & 69.8 & \textbf{23.8} & 82.8$^\dagger$ \\
Qwen2.5-VL-7B & Qwen-LookAgain~\cite{qwen-lookagain} & 69.8$^\dagger$ & 26.4 & 37.8$^\dagger$ & 82.8 & 65.9 & 69.2 & 22.6 & 81.4$^\dagger$ \\
\rowcolor{regroundpurple}
Qwen2.5-VL-7B & \textbf{ReGround} & \textbf{72.0} & \textbf{29.8} & \textbf{40.0} & \underline{83.8} & \textbf{67.6} & \textbf{71.5} & \underline{23.6} & \underline{83.2} \\
\midrule
Qwen3-VL-8B & Base & 77.2 & 46.7 & 45.5 & 84.5 & 70.9 & 74.6 & 22.5 & 90.2 \\
\rowcolor{regroundpurple}
Qwen3-VL-8B & \textbf{ReGround} & \textbf{79.8} & \textbf{50.1}
& \textbf{48.3} & \textbf{85.8} & \textbf{72.6} & \textbf{77.3}
& \textbf{25.9} & \textbf{91.3} \\
\bottomrule
\end{tabular}
\vspace{1pt}
{\footnotesize
$^\dagger$Reproduced under VLMEvalKit. Unmarked numbers from original papers.
}
\end{table*}

By appropriately tuning these coefficients, Equation~\ref{eq:reward} elegantly collapses into an asymmetric four-quadrant distribution (assuming correct formatting, thus $\lambda_{form} R_{form} = 0$):
\begin{itemize}
    \item \textbf{Highest Reward (Recovery/Verification):} Structurally valid re-grounding ($\mathbb{I}_{reg}\!=\!1$) and correct ($\mathbb{I}_{acc}\!=\!1$).
    \item \textbf{Moderate Reward (Efficient Precision):} No re-grounding ($\mathbb{I}_{reg}\!=\!0$) but correct ($\mathbb{I}_{acc}\!=\!1$).
    \item \textbf{Partial Reward (Attempted Correction):} Structurally valid re-grounding ($\mathbb{I}_{reg}\!=\!1$) but incorrect ($\mathbb{I}_{acc}\!=\!0$).
    \item \textbf{Penalty (Blind Failure):} No re-grounding ($\mathbb{I}_{reg}\!=\!0$) and incorrect ($\mathbb{I}_{acc}\!=\!0$).
\end{itemize}

After reinforcement learning, we obtain the updated policy $\pi_{\text{RL}}$, which exhibits improved structural compliance, reflection consistency, and answer accuracy.
This intentional asymmetry serves two critical purposes.

\textbf{(a) Behavioral survival.} The partial reward keeps re-grounding behavior from vanishing early in optimization, sustaining exploration until it yields accuracy gains.
\textbf{(b) Discouraging unverified guessing.} It mitigates the risk of models exploiting language priors to guess answers without visual grounding. Concurrently, since successful recovery (\emph{reground+correct}) inherently dominates failed attempts (\emph{reground+incorrect}), GRPO ensures the policy ultimately converges toward actual accuracy rather than mere verbosity.
We employ standard GRPO without algorithmic modifications, explicitly concentrating our technical contribution on data curriculum and reward shaping. This design enables seamless reproduction via off-the-shelf frameworks (e.g., LLaMA-Factory~\cite{llamafactory} and VeRL~\cite{verl}).


\section{Experiments}
\label{sec:experiments}

We evaluate ReGround on eight multimodal benchmarks with three goals:
to assess its overall effectiveness and cost (\S\ref{sec:main_results}),
to identify which components drive the gains (\S\ref{sec:why}),
and to test whether re-examination redirects attention toward diagnosed regions (\S\ref{sec:visual_evidence}).

\subsection{Experimental Setup}
\label{sec:setup}

\noindent\textbf{Benchmarks.}
MathVista~\cite{mathvista}, MathVision~\cite{mathvision},
MathVerse~\cite{mathverse} (visual-only),
HallusionBench~\cite{hallusionbench},
MMBench~\cite{mmbench}, MMStar~\cite{mmstar},
VisuLogic~\cite{visulogic}, V*Bench~\cite{vstarbench}.

\noindent\textbf{Backbones.}
Qwen2.5-VL-7B~\cite{qwen25vl} (primary) and Qwen3-VL-8B~\cite{qwen3vl} (cross-backbone verification).

\noindent\textbf{Baselines.}
Thyme~\cite{thyme} (tool-augmented),
Look-Back~\cite{look-back} (implicit re-focusing),
Qwen-LookAgain~\cite{qwen-lookagain} (custom RL),
R1-VL~\cite{r1-vl}.
$^\dagger$: reproduced under VLMEvalKit.

\noindent\textbf{Training.}
We first perform supervised fine-tuning on 68,477 trajectories constructed from ten source datasets, then apply GRPO for policy refinement.
Unless otherwise specified, all ablations use Qwen2.5-VL-7B.
Detailed hyperparameters and compute budget are provided in the supplementary material.

\subsection{Main Results}
\label{sec:main_results}

On Qwen2.5-VL-7B (Table~\ref{tab:main_results}), ReGround attains the best performance on five of eight benchmarks, including all three math-reasoning benchmarks (MathVista +3.8, MathVision +2.8, MathVerse +4.5), and remains within 0.3 points of the strongest baseline on the remaining three.
The gains are particularly pronounced on HallusionBench (+6.8), which specifically targets visual hallucination---the failure mode ReGround is designed to address.
On Qwen3-VL-8B, the same training recipe yields gains across all eight benchmarks (+1.1 to +3.4) without hyperparameter re-tuning, suggesting cross-backbone transfer rather than scale invariance, though we note that baselines were not re-trained on this backbone.
The three benchmarks where ReGround does not lead---MMBench, VisuLogic, and V*Bench---exhibit a common characteristic: the task-critical visual information is often fine-grained or spatially distributed (e.g., logical arrangement in VisuLogic, small visual targets in V*Bench), where a single diagnostic cue may be insufficient to redirect attention comprehensively. In these cases, tool-augmented methods that can crop or zoom into sub-regions (Thyme) or implicit re-focusing that operates at the token level (Look-Back) may offer complementary advantages.

\paragraph{Efficiency.}
\label{sec:cost_analysis}
ReGround's gains come with moderate overhead (Table~\ref{tab:cost_overall}): 1.62$\times$ total tokens and 1.76$\times$ latency relative to the baseline, while improving accuracy by 4.9 points.
Compared with Thyme, ReGround uses slightly fewer tokens (1889 vs.\ 1935), runs substantially faster (1.39s vs.\ 2.41s), and achieves higher accuracy.
We attribute this difference to the fact that ReGround adds a conditional image re-injection step, whereas Thyme incurs longer sequential autoregressive generation through tool-use traces.
Because re-injection is conditional, non-triggered and triggered samples average 0.95s and 2.31s, respectively, yielding the reported 1.39s overall latency on this 2{,}255-sample subset.
Re-injection adds about 518 prompt-side tokens on average in this analysis, and its context cost grows roughly linearly for multi-image and video inputs; cue-guided region or token re-injection is a promising direction for reducing this overhead.
Detailed token-channel decomposition is provided in the supplementary material.

\begin{table}[t]
\centering
\small
\caption{Efficiency--accuracy trade-off (sample-count-weighted avg.\ over MathVista, MathVision, HallusionBench; 2,255 samples).}
\label{tab:cost_overall}
\setlength{\tabcolsep}{5pt}
\begin{tabular}{lcccc}
\toprule
\textbf{Method} & \textbf{Acc.} & \textbf{Total Tok.} & \textbf{Calls} & \textbf{Time} \\
\midrule
Baseline & 61.2 & 1163\,(1.00$\times$) & 1.00 & 0.79s\,(1.00$\times$) \\
Thyme    & 64.7 & 1935\,(1.66$\times$) & 1.10 & 2.41s\,(3.05$\times$) \\
\rowcolor{regroundpurple}
\textbf{ReGround} & \textbf{66.1} & 1889\,(1.62$\times$) & 1.34 & 1.39s\,(1.76$\times$) \\
\bottomrule
\end{tabular}
\vspace{-4pt}
\end{table}

\subsection{Why Does ReGround Work?}
\label{sec:why}

\subsubsection{Is it just better data or text reflection?}
\label{sec:core_ablation}

\begin{table}[t]
\centering
\small
\caption{Core mechanism ablation.}
\label{tab:core_ablation}
\setlength{\tabcolsep}{5pt}
\begin{tabular}{clccc}
\toprule
\textbf{\#} & \textbf{Setting} & \textbf{MathVista} & \textbf{MathVision} & \textbf{Hallus.} \\
\midrule
\ding{172} & Baseline & 68.2 & 27.0 & 64.7 \\
\ding{173} & + CoT-only SFT & 68.8\,{\scriptsize(+0.6)} & 27.4\,{\scriptsize(+0.4)} & 66.1\,{\scriptsize(+1.4)} \\
\ding{174} & + Text-only R2 SFT & 69.1\,{\scriptsize(+0.9)} & 27.6\,{\scriptsize(+0.6)} & 66.8\,{\scriptsize(+2.1)} \\
\ding{175} & + Visual R2 SFT & 70.6\,{\scriptsize(+2.4)} & 28.7\,{\scriptsize(+1.7)} & 69.7\,{\scriptsize(+5.0)} \\
\midrule
\ding{176} & + Text-only R2 + GRPO & 69.9\,{\scriptsize(+1.7)} & 28.2\,{\scriptsize(+1.2)} & 68.3\,{\scriptsize(+3.6)} \\
\rowcolor{regroundpurple}
\ding{177} & + Visual R2 + GRPO (\textbf{Full}) & \textbf{72.0}\,{\scriptsize(+3.8)} & \textbf{29.8}\,{\scriptsize(+2.8)} & \textbf{71.5}\,{\scriptsize(+6.8)} \\
\bottomrule
\end{tabular}
\vspace{-4pt}
\end{table}

Table~\ref{tab:core_ablation} disentangles the contributions of trajectory supervision, textual reflection, visual re-injection, and RL refinement.

\noindent\textbf{It is not just better SFT data.}
Flattening the same 68K trajectories into single-round CoT (\ding{173}) yields only marginal gains (+0.6 on MathVista), far below visual R2 SFT (\ding{175}: +2.4), indicating that the benefit does not come from data scale or answer quality alone.

\noindent\textbf{It is not just text reflection.}
Replacing image re-injection with text-only self-reflection (\ding{174}) improves over the baseline but remains clearly below visual R2, especially on HallusionBench (+2.1 vs.\ +5.0), confirming that visual re-injection (\S\ref{sec:reinjection}) is essential and cannot be substituted by textual re-reading alone.
This gap is consistent with our motivation (Figure~\ref{fig:motivation}): once visual information has decayed, textual self-reflection cannot recover details that the model no longer attends to.

\noindent\textbf{GRPO further improves, but does not replace, visual grounding.}
RL refinement boosts the visual R2 model by another +1.4 (MathVista) and +1.8 (HallusionBench), while text-only R2 + GRPO (\ding{176}) still falls below visual R2 SFT (\ding{175}).
This result is noteworthy: even with a well-calibrated triggering policy, RL cannot compensate for the absence of visual input in Round~2, confirming that the re-injection mechanism---not just the training recipe---is essential.

\subsubsection{Does each trajectory mode matter?}
\label{sec:data_ablation}

\begin{table}[!t]
\centering
\small
\caption{Data composition ablation (Stage~1 SFT only). Trigger: \% of samples invoking \texttt{<reground>}.}
\label{tab:data_ablation}
\setlength{\tabcolsep}{4pt}
\begin{tabular}{lcccc}
\toprule
\textbf{SFT Data} & \textbf{MathVista} & \textbf{MathVision} & \textbf{Hallus.} & \textbf{Trigger} \\
\midrule
Full & 70.6 & 28.7 & 69.7 & $\sim$32\% \\
w/o No-ReGround & 69.5\,{\scriptsize(-1.1)} & 27.9\,{\scriptsize(-0.8)} & 67.2\,{\scriptsize(-2.5)} & $\sim$85\% \\
w/o Grounding (Step~2) & 69.1\,{\scriptsize(-1.5)} & 27.5\,{\scriptsize(-1.2)} & 66.8\,{\scriptsize(-2.9)} & $\sim$22\% \\
w/o Verification (Step~3) & 69.8\,{\scriptsize(-0.8)} & 28.1\,{\scriptsize(-0.6)} & 68.9\,{\scriptsize(-0.8)} & $\sim$25\% \\
\bottomrule
\end{tabular}
\vspace{-4pt}
\end{table}

Table~\ref{tab:data_ablation} examines how different trajectory modes contribute to the Stage~1 training signal.

\noindent\textbf{Grounding trajectories (Step~2) are the most impactful.}
Removing Grounding Information Re-injection trajectories causes the largest accuracy drop across all benchmarks (HallusionBench: $-2.9$), confirming that ``correct but ungrounded'' samples---where the diagnostic cue forces the model to anchor speculative reasoning to concrete visual elements (\S\ref{sec:reinjection})---are essential for teaching the model that re-examination is valuable even when the initial answer appears correct.

\noindent\textbf{No-ReGround prevents over-triggering.}
Without negative examples, the trigger rate spikes from $\sim$32\% to $\sim$85\%, degrading accuracy---especially on HallusionBench ($-2.5$), where unnecessary re-examination perturbs already-correct answers.
This result underscores the importance of balanced training data: the model must learn not only \emph{when} to re-examine but also \emph{when to refrain}.

\noindent\textbf{Verification (Step~3) provides supplementary value.}
The per-benchmark drop from removing Verification Information Re-injection trajectories is the smallest ($-0.8$ MathVista), but the reduced trigger rate ($32\%\to25\%$) suggests that this confirmation-oriented mode (\S\ref{sec:reinjection}) helps maintain a less conservative triggering policy.

\noindent\textbf{Reward calibration confirms asymmetric risk.}
We further sweep an exploration-preference (EP) index that controls the reward bias toward re-examination (full results in the supplementary material).
When the reward becomes too conservative, the trigger rate collapses and the correction benefit is under-exploited; when it becomes too aggressive, the trigger rate rises sharply and accuracy degrades through excessive negative flips.
The best overall setting emerges at an intermediate trigger rate (about 45\%), confirming that conservative triggering forgoes recoverable gains, whereas over-triggering actively harms accuracy.

\begin{figure*}[t]
\centering
\includegraphics[height=5.7cm]{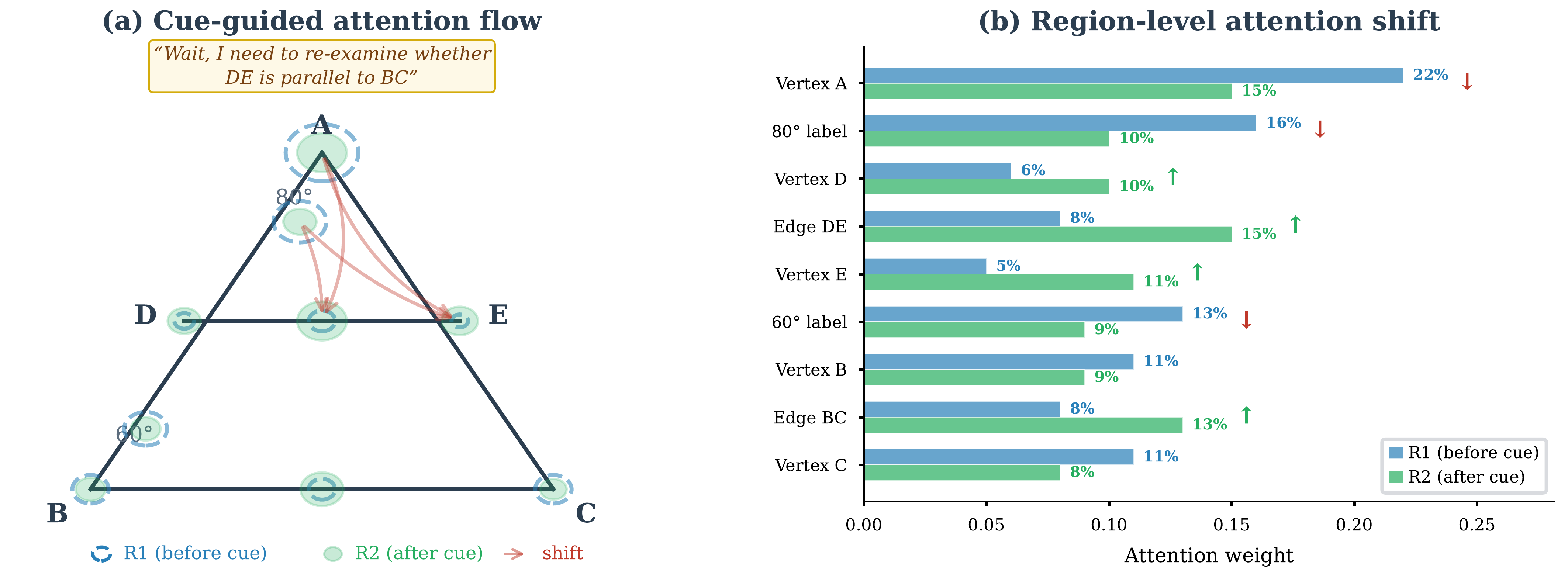}
\caption{Cue-guided attention redirection on a MathVerse geometry problem (Layer~20, Qwen2.5-VL-7B).
\textbf{Left:}~R1 attention (blue dashed) concentrates on Vertex~A and the 80\textdegree{} label; R2 attention (green filled) shifts after cue-guided re-examination.
\textbf{Right:}~Attention drains from apex regions and flows to cue-referenced segments: Edge~DE (+7pp), Vertex~E (+6pp), Edge~BC (+5pp).}
\Description{Left: geometric diagram with overlaid attention circles showing R1 concentration at apex and R2 shift to parallel segments. Right: horizontal bar chart comparing per-region attention weights before and after re-injection.}
\label{fig:attention}
\end{figure*}

\subsubsection{Does diagnostic quality matter?}
\label{sec:cue_ablation}

\begin{table}[!t]
\centering
\small
\caption{Cue quality ablation. Triggering held constant; only diagnostic content varies. MathVi.\,=\,MathVista, MathVe.\,=\,MathVerse, MathVn.\,=\,MathVision.}
\label{tab:cue_ablation}
\setlength{\tabcolsep}{4.5pt}
\begin{tabular}{lcccc}
\toprule
\textbf{Cue Source} & \textbf{MathVi.} & \textbf{MathVe.} & \textbf{MathVn.} & \textbf{Hallus.} \\
\midrule
No-ReGround & 68.2 & 35.5 & 27.0 & 64.7 \\
Generic$^{*}$ & 67.4\,{\scriptsize(\textcolor{red}{$-$0.8})} & 36.1\,{\scriptsize(+0.6)} & 27.2\,{\scriptsize(+0.2)} & 62.6\,{\scriptsize(\textcolor{red}{$-$2.1})} \\
Oracle (72B)$^{\dagger}$ & \textbf{72.4}\,{\scriptsize(+4.2)} & \textbf{41.2}\,{\scriptsize(+5.7)} & \textbf{30.5}\,{\scriptsize(+3.5)} & \textbf{73.3}\,{\scriptsize(+8.6)} \\
\rowcolor{regroundpurple}
Self-generated & 72.0\,{\scriptsize(+3.8)} & 40.0\,{\scriptsize(+4.5)} & 29.8\,{\scriptsize(+2.8)} & 71.5\,{\scriptsize(+6.8)} \\
\midrule
\textit{Recovery (Self\,/\,Oracle)} & 90\% & 79\% & 80\% & 79\% \\
\bottomrule
\end{tabular}
\vspace{1pt}
{\footnotesize $^{*}$``Please look at the image again carefully.'' \\
$^{\dagger}$The 72B variant shares the same vision encoder architecture as the 7B model.\\
Recovery $=$ (Self $-$ No-ReGround)\,/\,(Oracle $-$ No-ReGround).}
\vspace{-4pt}
\end{table}

To isolate cue quality from trigger timing, we hold trigger positions fixed and vary only the content inside \texttt{<reground>}.
Table~\ref{tab:cue_ablation} reveals a clear three-tier pattern.

\noindent\textbf{Without diagnosis, re-examination can be harmful.}
The generic prompt drops MathVista by $-0.8$ and HallusionBench by $-2.1$ below the no-re-examination baseline---undirected second-guessing overrides correct first-round answers.
This confirms that the benefit of re-examination is not automatic: image re-injection must be paired with a meaningful diagnostic signal.

\noindent\textbf{Diagnostic quality is a major determinant.}
The gap between generic and targeted cues reaches 4.6 points on MathVista and 8.9 on HallusionBench, indicating that \emph{what} the model re-examines strongly affects whether re-examination helps or hurts.
This finding is the empirical basis for our central claim: diagnostic quality---not the re-examination format alone---is the key variable governing self-correction effectiveness.

\noindent\textbf{Self-diagnosis recovers 79--90\% of oracle capacity.}
After SFT, the policy model generates cues that recover 79--90\% of the oracle variant's gains across all four benchmarks.
Since both variants share the same vision encoder architecture, the recovery rate primarily reflects learned diagnostic reasoning rather than visual perception differences.
This demonstrates that capability bootstrapping successfully transfers the diagnostic skill: the stronger variant is needed only for data construction and can be discarded at inference time.
The recovery pattern is also informative: transfer is strongest on MathVista (90\%) and remains solid on MathVision (80\%), while the residual gap is larger on MathVerse and HallusionBench (both 79\%).
This suggests that diagnostic transfer is more complete on domains closer to the SFT training distribution, whereas broader visual-semantic coverage may still help on more diverse grounding failures. We therefore treat the oracle only as a diagnostic upper bound.

\subsection{Visual Evidence of Cue-Guided Re-Grounding}
\label{sec:visual_evidence}

We emphasize that attention entropy measures the breadth of attention coverage rather than grounding quality directly.
We therefore interpret the following redistribution patterns as proxy evidence that image re-injection changes the attention distribution, and use SAS to assess whether the shift is task-relevant.

Figure~\ref{fig:attention} visualizes how the self-generated diagnostic cue redirects attention on a representative geometry problem.
After Round~1, attention concentrates on the apex region (Vertex~A: 22\%).
After image re-injection with the cue \emph{``...whether DE is parallel to BC''}, attention shifts primarily toward the cue-referenced segments---the regions required to verify parallelism.
This redistribution is not random: the regions gaining attention (Edge~DE, Vertex~E, Edge~BC) are precisely the elements mentioned in the diagnostic cue, while regions losing attention (Vertex~A, 80\textdegree{} label) are those already well-represented in Round~1.

To move beyond a single qualitative example, we also measure a Semantic Alignment Score (SAS).
SAS quantifies the change in attention mass on cue-referenced image regions from Round~1 to Round~2, where cue-referenced regions are identified via cross-attention from cue entity tokens to image tokens.
Across 2{,}510 triggered samples, 83\% show positive semantic alignment, with an average shift of +4.7 percentage points toward diagnosed regions.
Full entropy statistics, SAS methodology, and additional qualitative examples are provided in the supplementary material.

Attention analysis is secondary to behavioral evidence: under matched training, text-only R2 gains only +0.9, +0.6, and +2.1 points on MathVista, MathVision, and HallusionBench, versus +2.4, +1.7, and +5.0 for visual R2; even text-only R2 with GRPO trails visual R2 SFT (Table~\ref{tab:core_ablation}). Thus entropy and SAS are mechanism signatures, not direct evidence of accuracy.

\noindent\textbf{Joint interpretation.}
Together, these results separate renewed access from useful redirection. Re-injection makes visual tokens available again, but the diagnostic cue determines how that access is used. In Figure~\ref{fig:attention}, attention shifts toward the segments needed to verify parallelism. While this alignment is not causal proof, the text-only and cue-quality ablations show that targeted diagnosis is what turns renewed access into reliable recovery.

\noindent\textbf{Scope and limitations.}
ReGround targets localized grounding drift, not visual search or fine-grained localization; crop/zoom and token-level re-focusing remain complementary on spatially diffuse tasks. Re-injection adds about 518 prompt-side tokens per image and scales roughly linearly to multi-image or video inputs. Selective cue-guided region or token re-injection may reduce this cost. Re-injection cannot recover evidence lost during visual encoding.

\section{Conclusion}
\label{sec:conclusion}

We identify visual grounding decay as a persistent failure mode in multi-step VLM reasoning, and show that revisiting an image alone does not guarantee reliable self-correction.
Effective re-examination depends on \emph{what} to re-check: targeted self-diagnosis redirects attention to task-critical evidence, whereas unguided revisiting can induce second-guessing and erode accuracy.
\textbf{ReGround} combines capability-bootstrapped supervision with policy optimization to learn both \emph{when} to re-examine and \emph{what} to revisit, without external tools or architectural changes.
Across two VLM backbones and eight benchmarks, it improves accuracy with only moderate inference overhead.
These results establish diagnostic quality---not renewed visual access alone---as a central factor in visual self-correction: gains arise from revisiting evidence with the right intent.
As a complementary, tool-free mechanism, ReGround further motivates diagnosis-aware training for grounded multimodal reasoning.

\begin{acks}
This work was supported by the National Natural Science Foundation of China under Grant 42550106.
\end{acks}

\bibliographystyle{ACM-Reference-Format}
\bibliography{references}

\clearpage
\appendix

\section{Training Configuration}
\label{app:training}

\subsection{Stage~1: Supervised Fine-Tuning}

Table~\ref{tab:sft_params} summarizes the Stage~1 hyperparameters.
We use LLaMA-Factory with DeepSpeed ZeRO-3 for memory-efficient full-parameter fine-tuning.
Training runs for 2.5 epochs; the final checkpoint is selected at epoch~2.0 based on validation loss.
Only trajectories whose final answer matches the ground truth are retained.
Stage~1 SFT takes approximately 120 A100-GPU-hours; Stage~2 GRPO takes approximately 1920 A100-GPU-hours (including rollout generation).

\begin{table}[t]
\centering
\small
\caption{Stage~1 (SFT) hyperparameters.}
\label{tab:sft_params}
\setlength{\tabcolsep}{5pt}
\begin{tabular}{ll}
\toprule
\textbf{Hyperparameter} & \textbf{Value} \\
\midrule
Framework            & LLaMA-Factory \\
Backbone             & Qwen2.5-VL-7B-Instruct \\
Precision / Hardware & bf16 / 32$\times$A100-80G \\
Parallelism          & DeepSpeed ZeRO-3 \\
\midrule
Learning rate        & $8 \times 10^{-6}$ (Cosine) \\
Warmup ratio         & 0.05 \\
Weight decay / Grad norm & 0.1 / 1.0 \\
\midrule
Training data        & 68{,}477 trajectories \\
Epochs               & 2.5 (ckpt at 2.0)$^{\dagger}$ \\
Max sequence length  & 8{,}192 \\
Batch (per-device / accum) & 1 / 6 \\
\midrule
Stochastic trigger $\varepsilon$ & 0.15 \\
\bottomrule
\end{tabular}
\vspace{1pt}
{\footnotesize $^\dagger$Early stopping based on validation loss.}
\end{table}

\subsection{Stage~2: GRPO}

Table~\ref{tab:grpo_params} summarizes the Stage~2 hyperparameters.
We use VeRL with GRPO.
The reward yields four-quadrant values:
R+C $= 1.0$, D+C $= 0.6$, R+I $= 0.4$, D+I $= -0.1$.

\begin{table}[t]
\centering
\small
\caption{Stage~2 (GRPO) hyperparameters.}
\label{tab:grpo_params}
\setlength{\tabcolsep}{5pt}
\begin{tabular}{ll}
\toprule
\textbf{Hyperparameter} & \textbf{Value} \\
\midrule
Framework            & VeRL \\
Backbone             & Qwen2.5-VL-7B-Instruct \\
Precision / Hardware & bf16 / 32$\times$A100-80G \\
\midrule
Learning rate        & $1 \times 10^{-6}$ \\
Training steps       & 584 \\
Max response length  & 1{,}024 \\
\midrule
Group size $G$       & 8 \\
KL coefficient       & 0.01 \\
Clip ratio           & 0.2 \\
Rollout temperature  & 0.7 \\
\midrule
$\lambda_{reg}$ / $\lambda_{acc}$ / $\lambda_{form}$ & 0.5 / 0.7 / 0.01 \\
$\gamma$ / $\beta$   & 0.14 / 0.14 \\
\bottomrule
\end{tabular}
\end{table}

\subsection{Training Dynamics}

\begin{figure*}[t]
\centering
\includegraphics[width=\textwidth]{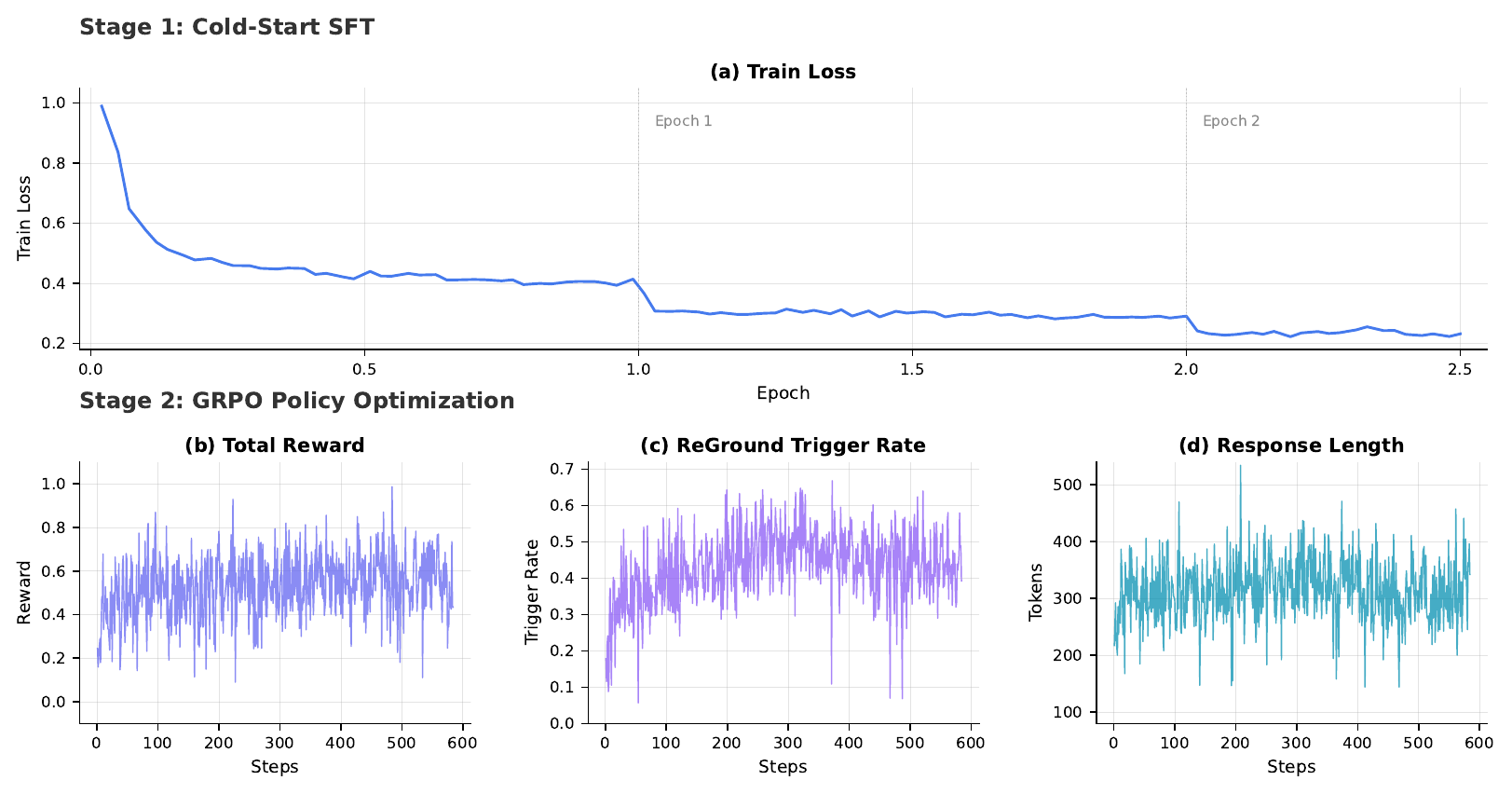}
\caption{Training dynamics.
\textbf{(a)}~SFT loss over 2.5 epochs (checkpoint at 2.0).
\textbf{(b)}~GRPO reward fluctuates 0.2--0.8 (discrete four-quadrant structure).
\textbf{(c)}~Trigger rate rises to 30--60\% within $\sim$50 steps, then stabilizes at 30--50\%.
\textbf{(d)}~Response length stays within 200--450 tokens.}
\label{fig:training_curves}
\end{figure*}

Figure~\ref{fig:training_curves} shows training curves for both stages.
The Stage~2 reward (b) lacks a smooth upward trajectory, reflecting the discrete four-quadrant reward where batch means depend on outcome-type proportions.
The trigger rate (c) better captures the training signal: rapid initial rise shows the policy learning to invoke \texttt{<reground>}, with subsequent stabilization around 40\% matching the optimal EP sweep setting (Table~\ref{tab:ep_sweep}).

\section{Data Construction Details}
\label{app:data}

\subsection{Source Datasets and Routing}

Table~\ref{tab:data_sources} summarizes the source datasets and trajectory routing distribution.
Among correct Round~1 responses (54{,}371), 22.9\% are classified as visually ungrounded (Step~2).
Among grounded correct responses (41{,}908), $\varepsilon = 0.15$ routes 19.4\% into Step~3.

\begin{table*}[t]
\centering
\small
\caption{Source datasets (left) and trajectory routing distribution (right) for Stage~1 construction.}
\label{tab:data_sources}
\begin{minipage}[t]{0.55\textwidth}
\vspace{0pt}%
\centering
\setlength{\tabcolsep}{3.5pt}
\begin{tabular}{llrc}
\toprule
\textbf{Dataset} & \textbf{Task Type} & \textbf{N} & \textbf{\%} \\
\midrule
CLEVR-Math  & Spatial reasoning      & 9{,}804 & 14.3 \\
IconQA      & Visual QA              & 8{,}761 & 12.8 \\
A-OKVQA     & Knowledge VQA          & 7{,}182 & 10.5 \\
DocQA       & Document understanding & 6{,}312 &  9.2 \\
DVQA        & Chart/diagram QA       & 6{,}127 &  8.9 \\
ScienceQA   & Science reasoning      & 5{,}898 &  8.6 \\
GeoQA       & Geometry               & 3{,}755 &  5.5 \\
Geometry3K  & Geometry               & 1{,}836 &  2.7 \\
TheoremQA   & Theorem proving        &    473  &  0.7 \\
Thyme (sub.)& Multi-step visual      & 18{,}329 & 26.8 \\
\midrule
\textbf{Total} &                     & \textbf{68{,}477} & \textbf{100} \\
\bottomrule
\end{tabular}
\end{minipage}%
\hfill%
\begin{minipage}[t]{0.38\textwidth}
\vspace{0pt}%
\centering
\setlength{\tabcolsep}{4pt}
\begin{tabular}{lrc}
\toprule
\textbf{Category} & \textbf{Count} & \textbf{\%} \\
\midrule
Step~1: Correction          & 14{,}106 & 20.6 \\
Step~2: Grounding           & 12{,}463 & 18.2 \\
Step~3: Verification        &  8{,}149 & 11.9 \\
No-ReGround                 & 33{,}759 & 49.3 \\
\midrule
\textbf{Total}              & \textbf{68{,}477} & \textbf{100} \\
\bottomrule
\end{tabular}
\end{minipage}
\end{table*}

\subsection{Prompt Templates}

\begin{figure*}[t]
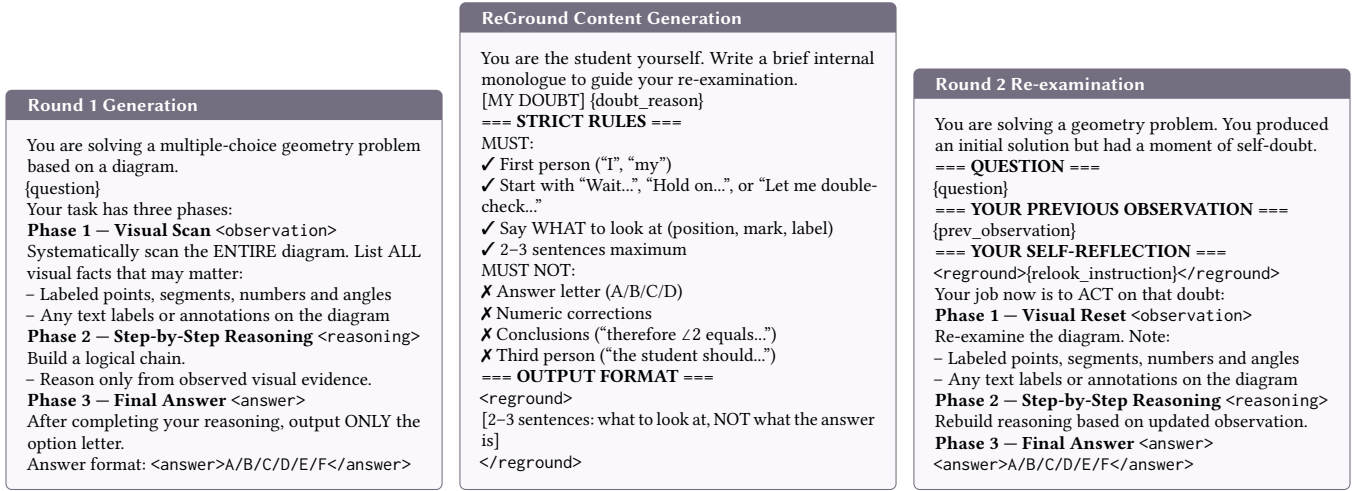

\centering

\begin{minipage}[t]{0.325\textwidth}
\begin{promptbox}[Round~1 Generation]{pipelinepurple}
You are solving a multiple-choice geometry problem based on a diagram.\\
\{question\}

Your task has three phases:

\textbf{Phase 1 --- Visual Scan} \texttt{<observation>}\\
Systematically scan the ENTIRE diagram.
List ALL visual facts that may matter:\\
-- Labeled points, segments, numbers and angles\\
-- Any text labels or annotations on the diagram

\textbf{Phase 2 --- Step-by-Step Reasoning} \texttt{<reasoning>}\\
Build a logical chain.\\
-- Reason only from observed visual evidence.

\textbf{Phase 3 --- Final Answer} \texttt{<answer>}\\
After completing your reasoning, output ONLY the option letter.\\
Answer format: \texttt{<answer>A/B/C/D/E/F</answer>}
\end{promptbox}
\end{minipage}
\hfill
\begin{minipage}[t]{0.325\textwidth}
\begin{promptbox}[ReGround Content Generation]{pipelinepurple}
You are the student yourself.
Write a brief internal monologue to guide your re-examination.

[MY DOUBT] \{doubt\_reason\}

\textbf{=== STRICT RULES ===}\\
MUST:\\
\ding{51} First person (``I'', ``my'')\\
\ding{51} Start with ``Wait...'', ``Hold on...'', or ``Let me double-check...''\\
\ding{51} Say WHAT to look at (position, mark, label)\\
\ding{51} 2--3 sentences maximum

MUST NOT:\\
\ding{55} Answer letter (A/B/C/D)\\
\ding{55} Numeric corrections\\
\ding{55} Conclusions (``therefore $\angle 2$ equals...'')\\
\ding{55} Third person (``the student should...'')

\textbf{=== OUTPUT FORMAT ===}\\
\texttt{<reground>}\\
{[2--3 sentences: what to look at, NOT what the answer is]}\\
\texttt{</reground>}
\end{promptbox}
\end{minipage}
\hfill
\begin{minipage}[t]{0.325\textwidth}
\begin{promptbox}[Round~2 Re-examination]{pipelinepurple}
You are solving a geometry problem.
You produced an initial solution but had a moment of self-doubt.

\textbf{=== QUESTION ===}\\
\{question\}

\textbf{=== YOUR PREVIOUS OBSERVATION ===}\\
\{prev\_observation\}

\textbf{=== YOUR SELF-REFLECTION ===}\\
\texttt{<reground>}\{relook\_instruction\}\texttt{</reground>}

Your job now is to ACT on that doubt:

\textbf{Phase 1 --- Visual Reset} \texttt{<observation>}\\
Re-examine the diagram. Note:\\
-- Labeled points, segments, numbers and angles\\
-- Any text labels or annotations on the diagram

\textbf{Phase 2 --- Step-by-Step Reasoning} \texttt{<reasoning>}\\
Rebuild reasoning based on updated observation.

\textbf{Phase 3 --- Final Answer} \texttt{<answer>}\\
\texttt{<answer>A/B/C/D/E/F</answer>}
\end{promptbox}
\end{minipage}

\vspace{4pt}
\caption{Reasoning pipeline prompt templates executed by the policy model.
\textbf{(a)}~Round~1 produces initial observation, reasoning, and answer.
\textbf{(b)}~ReGround content generation converts a diagnostic doubt into a first-person monologue without revealing the answer.
\textbf{(c)}~Round~2 re-examination rebuilds reasoning from updated visual observations.
Geometry variant shown; other domains use adapted terminology.}
\label{fig:prompt_pipeline}
\end{figure*}

\begin{figure*}[t]
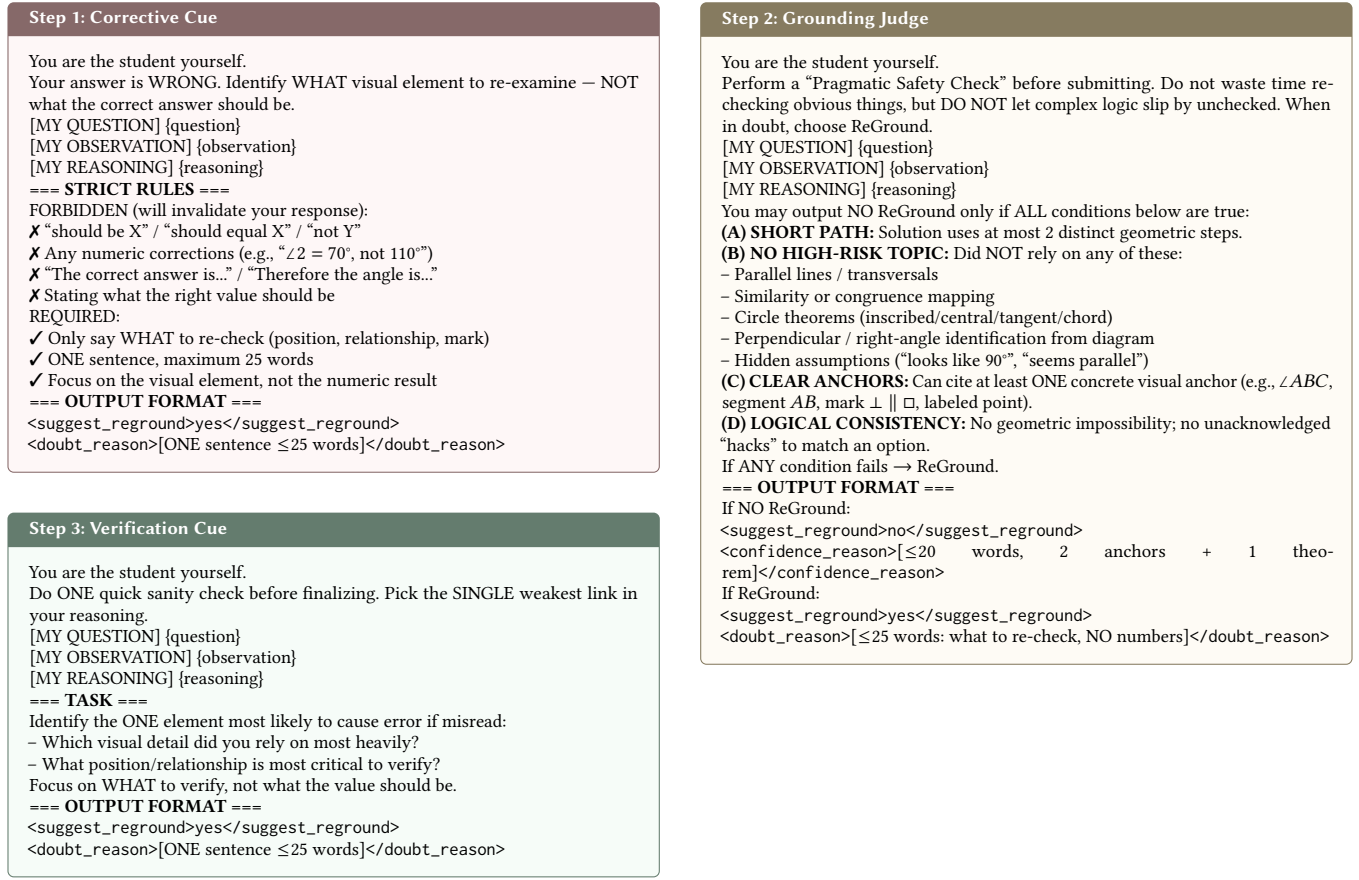

\centering
\noindent\begin{minipage}[t]{0.485\textwidth}%
\vspace{0pt}%
\begin{promptbox}[Step~1: Corrective Cue]{correctionred}
You are the student yourself.\\
Your answer is WRONG. Identify WHAT visual element to re-examine --- NOT what the correct answer should be.

[MY QUESTION] \{question\}\\
{[MY OBSERVATION]} \{observation\}\\
{[MY REASONING]} \{reasoning\}

\textbf{=== STRICT RULES ===}\\
FORBIDDEN (will invalidate your response):\\
\ding{55} ``should be X'' / ``should equal X'' / ``not Y''\\
\ding{55} Any numeric corrections (e.g., ``$\angle 2 = 70$\textdegree, not 110\textdegree'')\\
\ding{55} ``The correct answer is...'' / ``Therefore the angle is...''\\
\ding{55} Stating what the right value should be

REQUIRED:\\
\ding{51} Only say WHAT to re-check (position, relationship, mark)\\
\ding{51} ONE sentence, maximum 25 words\\
\ding{51} Focus on the visual element, not the numeric result

\textbf{=== OUTPUT FORMAT ===}\\
\texttt{<suggest\_reground>yes</suggest\_reground>}\\
\texttt{<doubt\_reason>}[ONE sentence $\leq$25 words]\texttt{</doubt\_reason>}
\end{promptbox}

\vspace{4pt}

\begin{promptbox}[Step~3: Verification Cue]{verificationgreen}
You are the student yourself.\\
Do ONE quick sanity check before finalizing.
Pick the SINGLE weakest link in your reasoning.

[MY QUESTION] \{question\}\\
{[MY OBSERVATION]} \{observation\}\\
{[MY REASONING]} \{reasoning\}

\textbf{=== TASK ===}\\
Identify the ONE element most likely to cause error if misread:\\
-- Which visual detail did you rely on most heavily?\\
-- What position/relationship is most critical to verify?

Focus on WHAT to verify, not what the value should be.

\textbf{=== OUTPUT FORMAT ===}\\
\texttt{<suggest\_reground>yes</suggest\_reground>}\\
\texttt{<doubt\_reason>}[ONE sentence $\leq$25 words]\texttt{</doubt\_reason>}
\end{promptbox}
\end{minipage}%
\hfill%
\begin{minipage}[t]{0.485\textwidth}%
\vspace{0pt}%
\begin{promptbox}[Step~2: Grounding Judge]{groundingyellow}
You are the student yourself.\\
Perform a ``Pragmatic Safety Check'' before submitting.
Do not waste time re-checking obvious things,
but DO NOT let complex logic slip by unchecked.
When in doubt, choose ReGround.

[MY QUESTION] \{question\}\\
{[MY OBSERVATION]} \{observation\}\\
{[MY REASONING]} \{reasoning\}

You may output NO ReGround only if ALL conditions below are true:\\
\textbf{(A) SHORT PATH:} Solution uses at most 2 distinct geometric steps.\\
\textbf{(B) NO HIGH-RISK TOPIC:} Did NOT rely on any of these:\\
\quad-- Parallel lines / transversals\\
\quad-- Similarity or congruence mapping\\
\quad-- Circle theorems (inscribed/central/tangent/chord)\\
\quad-- Perpendicular / right-angle identification from diagram\\
\quad-- Hidden assumptions (``looks like 90\textdegree'', ``seems parallel'')\\
\textbf{(C) CLEAR ANCHORS:} Can cite at least ONE concrete visual anchor
(e.g., $\angle ABC$, segment $AB$, mark $\perp$ $\parallel$ $\square$, labeled point).\\
\textbf{(D) LOGICAL CONSISTENCY:} No geometric impossibility;
no unacknowledged ``hacks'' to match an option.

If ANY condition fails $\to$ ReGround.

\textbf{=== OUTPUT FORMAT ===}\\
If NO ReGround:\\
\texttt{<suggest\_reground>no</suggest\_reground>}\\
\texttt{<confidence\_reason>}[$\leq$20 words, 2 anchors + 1 theorem]\texttt{</confidence\_reason>}

If ReGround:\\
\texttt{<suggest\_reground>yes</suggest\_reground>}\\
\texttt{<doubt\_reason>}[$\leq$25 words: what to re-check, NO numbers]\texttt{</doubt\_reason>}
\end{promptbox}
\end{minipage}

\vspace{4pt}
\caption{Trigger-condition prompt templates.
\textbf{Left-top:}~Corrective cue (Step~1) identifies what to re-examine when the answer is incorrect, with strict answer-leakage prevention.
\textbf{Left-bottom:}~Verification cue (Step~3) identifies the single weakest link for stochastic confirmation.
\textbf{Right:}~Grounding judge (Step~2) evaluates whether correct reasoning is visually grounded via four domain-specific criteria.
The \texttt{doubt\_reason} output feeds into Figure~\ref{fig:prompt_pipeline}(b).
Geometry variant shown; other domains use adapted criteria.
The four conditions shown here are the geometry-specific operationalization of the general grounding criteria defined in \S3.1.1 of the main text.}
\label{fig:prompt_triggers}
\end{figure*}

Figures~\ref{fig:prompt_pipeline} and~\ref{fig:prompt_triggers} show the six prompt templates used in trajectory construction.
All templates follow a shared structure but are domain-adapted (geometry variant shown).
Key design principles:
(1)~No answer leakage in diagnostic cues;
(2)~First-person framing for internalization;
(3)~Structured output schema for automated parsing.


\section{Reward Design: EP Sweep}
\label{app:reward}

The asymmetric reward is parameterized by
$[r(\text{R}\!\wedge\!\checkmark),\, r(\neg\!\text{R}\!\wedge\!\checkmark),\, r(\text{R}\!\wedge\!\times),\, r(\neg\!\text{R}\!\wedge\!\times)]$.
We fix $r(\text{R}\!\wedge\!\checkmark)=1.0$, $r(\neg\!\text{R}\!\wedge\!\times)=-0.1$,
and sweep EP from $-1.0$ to $+1.0$ (Table~\ref{tab:ep_sweep}).

\noindent\textbf{Both extremes fail.}
EP$=-1.0$: trigger rate 12\% (wasteful).
EP$=+1.0$: trigger rate 74\%, accuracy below baseline.

\noindent\textbf{EP$=+0.2$ is optimal} (45\% trigger rate).
Task-specific rates: MathVista 45\%, MathVerse 38\%, MathVision 26\%, HallusionBench 16\%.

\begin{table}[t]
\centering
\small
\caption{EP sweep. Best per-column in \textbf{bold}.}
\label{tab:ep_sweep}
\setlength{\tabcolsep}{3pt}
\begin{tabular}{rclcccc}
\toprule
\textbf{EP} & \textbf{Trig} & \textbf{Setting} & \textbf{MVi.} & \textbf{MVe.} & \textbf{MVn.} & \textbf{Hal.} \\
\midrule
--- & --- & Baseline & 68.2 & 35.5 & 27.0 & 64.7 \\
--- & --- & SFT Only & 70.6 & 38.3 & 28.7 & 69.7 \\
\midrule
$-1.0$ & 12\% & Corr.-Only & 69.3 & 37.0 & 27.5 & 66.3 \\
$-0.5$ & 19\% & NR+C & 70.8 & 38.1 & 28.1 & 69.1 \\
$0.0$  & 38\% & Balanced & 71.1 & 39.4 & \textbf{30.2} & 70.3 \\
\rowcolor{regroundpurple}
$+0.2$ & 45\% & \textbf{Ours} & \textbf{72.0} & \textbf{40.0} & 29.8 & \textbf{71.5} \\
$+0.5$ & 64\% & R+I & 69.0 & 37.7 & 26.6 & 68.1 \\
$+1.0$ & 74\% & RG-Only & 67.7 & 35.0 & 25.5 & 66.9 \\
\bottomrule
\end{tabular}
\end{table}

\subsection{Flip-Rate Decomposition}

Table~\ref{tab:fliprate} decomposes accuracy changes into positive flips (wrong$\to$correct) and negative flips (correct$\to$wrong).
Key observations:
(1)~posFlip saturates at $\sim$5.4\% beyond 44\% trigger rate;
(2)~negFlip rises sharply from 1.6\% to 5.8\%;
(3)~the best pf/nf ratio occurs at EP$=+0.2$ (3.4$\times$);
(4)~HallusionBench is more robust to over-triggering.

\begin{table*}[t]
\centering
\small
\caption{Flip-rate decomposition. pf: wrong$\to$correct; nf: correct$\to$wrong.}
\label{tab:fliprate}
\setlength{\tabcolsep}{4pt}
\begin{tabular}{l r cccc cccc}
\toprule
& & \multicolumn{4}{c}{\textbf{MathVista}} & \multicolumn{4}{c}{\textbf{HallusionBench}} \\
\cmidrule(lr){3-6} \cmidrule(lr){7-10}
\textbf{Setting} & \textbf{EP} & Trig & pf & nf & Acc & Trig & pf & nf & Acc \\
\midrule
Correctness-only & $-1.0$ & 11.5 & 1.6 & 0.5 & 69.3 & 3.4 & 1.8 & 0.2 & 66.3 \\
NR+C Heavy & $-0.5$ & 18.7 & 2.9 & 0.6 & 70.5 & 8.1 & 4.7 & 0.3 & 69.1 \\
SFT Only & --- & 32.4 & 3.8 & 1.4 & 70.6 & 13.6 & 5.8 & 0.8 & 69.7 \\
Balanced & $0.0$ & 38.4 & 4.2 & 1.9 & 70.5 & 15.2 & 6.7 & 1.1 & 70.3 \\
\rowcolor{regroundpurple}
\textbf{Ours} & $+0.2$ & 44.8 & 5.4 & 1.6 & \textbf{72.0} & 15.8 & 6.9 & 1.2 & \textbf{71.5} \\
R+I Heavy & $+0.5$ & 63.5 & 5.0 & 4.2 & 69.0 & 34.7 & 6.1 & 2.7 & 68.1 \\
RG-Only & $+1.0$ & 74.2 & 5.3 & 5.8 & 67.7 & 49.6 & 5.4 & 3.2 & 66.9 \\
\bottomrule
\end{tabular}
\end{table*}
\clearpage
\paragraph{Visual Attention Analysis.}
\label{app:attention}

We measure spatial entropy at Layer~20 of Qwen2.5-VL-7B at R1~Start, R1~End, and R2~Start (Table~\ref{tab:attention}).
Entropy measures attention \emph{breadth}, not grounding quality; we use the Semantic Alignment Score (SAS) to assess whether the observed attention redistribution is task-relevant.

\noindent\textbf{SAS} $= \frac{\sum_{i \in \mathcal{C}} a^{\mathrm{R2}}_i}{\sum_j a^{\mathrm{R2}}_j} - \frac{\sum_{i \in \mathcal{C}} a^{\mathrm{R1}}_i}{\sum_j a^{\mathrm{R1}}_j}$,
where $\mathcal{C}$ denotes the set of cue-referenced image tokens, identified as the top-$K{=}20$ tokens by cross-attention weight from diagnostic-cue entity tokens to image tokens.
A positive SAS indicates that the attention mass on cue-referenced regions increased from Round~1 to Round~2, consistent with the diagnostic cue guiding re-examination toward the intended visual elements.
Across 2{,}510 triggered samples from four benchmarks, 83\% show positive SAS with an average shift of +4.7 percentage points toward diagnosed regions.
Additionally, across the same 2{,}510 samples, entropy recovers
after image re-injection ($H_{\mathrm{R2S}} > H_{\mathrm{R1E}}$) in
94.4\% of cases, confirming that the decay-and-recovery pattern
reported in the main text is a robust benchmark-level phenomenon.
\begin{table}[!t]
\centering
\small
\caption{Attention redistribution (Layer~20).}
\label{tab:attention}
\setlength{\tabcolsep}{2.5pt}
\begin{tabular}{lr ccc cc cc}
\toprule
& & \multicolumn{5}{c}{\textbf{Entropy}} & \multicolumn{2}{c}{\textbf{Sem.~Align.}} \\
\cmidrule(lr){3-7} \cmidrule(lr){8-9}
\textbf{Bench.} & $n$ & $H_{\mathrm{R1S}}$ & $H_{\mathrm{R1E}}$ & $H_{\mathrm{R2S}}$ & $\Delta$ & \textbf{\%} & \textbf{SAS} & \textbf{\%$\uparrow$} \\
\midrule
MVista  & 539  & 6.12 & 4.36 & 5.72 & $-$1.76 & 98.9 & +5.1 & 86 \\
MVerse  & 1737 & 5.64 & 4.15 & 5.18 & $-$1.49 & 97.4 & +4.0 & 82 \\
MVision & 91   & 7.04 & 5.21 & 6.54 & $-$1.82 & 100  & +6.3 & 89 \\
Hallus. & 143  & 7.73 & 5.15 & 7.03 & $-$2.57 & 100  & +4.6 & 78 \\
\midrule
\textbf{All} & \textbf{2510} & \textbf{5.91} & \textbf{4.29} & \textbf{5.45} & $\mathbf{-}$\textbf{1.62} & \textbf{98.0} & \textbf{+4.7} & \textbf{83} \\
\bottomrule
\end{tabular}
\vspace{1pt}
{\footnotesize $\Delta = H_{\mathrm{R1E}} - H_{\mathrm{R1S}}$; \% = $\Delta < 0$; \%$\uparrow$ = SAS $> 0$.}
\end{table}

\paragraph{Detailed Efficiency Statistics.}
\label{app:efficiency}

ReGround pays more input cost (+518 tokens for re-injected image) but substantially less generation cost than Thyme (+208 vs.\ +392 output tokens), as shown in Table~\ref{tab:efficiency_detailed}.
The decomposition clarifies the efficiency difference: ReGround's overhead comes primarily from the prompt side (additional visual turn), while its generation cost remains moderate because the diagnostic cue and revised reasoning are typically concise.

\begin{table}[!t]
\centering
\small
\caption{Token-channel decomposition (2{,}255 samples).}
\label{tab:efficiency_detailed}
\setlength{\tabcolsep}{3.5pt}
\begin{tabular}{lcccccc}
\toprule
\textbf{Method} & \textbf{In} & \textbf{Out} & \textbf{Total} & $\Delta$In & $\Delta$Out & \textbf{Time} \\
\midrule
Baseline & 894 & 269 & 1163 & --- & --- & 0.79s \\
Thyme & 1274 & 661 & 1935 & +380 & +392 & 2.41s \\
\rowcolor{regroundpurple}
\textbf{RG} & 1412 & 476 & 1889 & +518 & +208 & 1.39s \\
\bottomrule
\end{tabular}
\end{table}

\paragraph{Grounding Judge Details.}
\label{app:grounding_judge}

The Step~2 judge uses a larger variant from the same family (Qwen2.5-VL-72B-Instruct) for stronger instruction-following.
It evaluates four criteria:
(1)~visual attribute coverage;
(2)~alternative elimination via visual evidence;
(3)~hedging absence;
(4)~visual anchor.
In practice, geometry prompts additionally assess solution complexity and high-risk reasoning patterns (e.g., parallel-line assumptions, circle theorems); chart/document prompts assess label verification and axis reading.
Across all 68{,}477 trajectories, 22.9\% of correct Round~1 responses are classified as visually ungrounded and routed to Step~2 re-examination.

\noindent\textbf{Limitation.}
Human-annotated agreement has not been measured.
The judge affects only training-data routing, not benchmark scoring, so potential bias impacts data composition rather than evaluation integrity.

\section{Qualitative Analysis}
\label{app:qualitative}

We present representative trajectories that illustrate the key behavioral patterns of ReGround.
Each example uses actual model outputs from Qwen2.5-VL-7B after Stage~2 training.
Taken together, the four figures below cover all trajectory types from the training curriculum: generic-vs-targeted cue comparison (Figure~\ref{fig:case_generic}), verification (Figure~\ref{fig:case_geometry}), correction (Figure~\ref{fig:case_correction}), and No-ReGround (Figure~\ref{fig:case_chartmed}).

\subsection{Generic vs.\ Targeted Diagnostic Cue}

\begin{figure*}[t]
\centering

\begin{tcolorbox}[colback=gray!8, colframe=gray!40, boxrule=0.4pt, arc=2pt, left=5pt, right=5pt, top=4pt, bottom=4pt]
\begin{minipage}[c]{0.64\textwidth}
\small
\textbf{Q:} Subtract all tiny yellow matte cylinders. Subtract all small blue rubber spheres. How many objects are left?
\end{minipage}
\hfill
\begin{minipage}[c]{0.28\textwidth}
\centering
\includegraphics[width=0.95\textwidth]{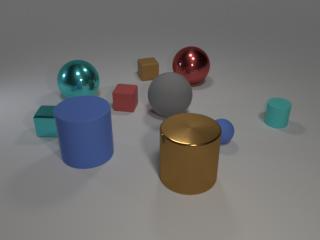}
\end{minipage}
\end{tcolorbox}

\vspace{3pt}

\begin{tcolorbox}[colback=gray!5, colframe=gray!30, boxrule=0.3pt, fontupper=\footnotesize, left=5pt, right=5pt, top=3pt, bottom=3pt]
\textbf{Round~1 (shared):}
\texttt{<think>} There are 10 objects in total in the image. Objects include: 2 large spheres (one teal metallic, one red metallic), 1 large gray matte sphere, 1 large blue matte cylinder, 1 large gold metallic cylinder, 1 small cyan matte cube, 1 small red matte cube, 1 small brown matte cube, 1 small blue matte sphere, 1 small cyan matte cylinder. No tiny yellow matte cylinders are visible. There is 1 small blue rubber sphere (matte finish). Total number of objects: 10. There are 0 tiny yellow matte cylinders $\to$ subtract 0. Subtract small blue rubber spheres: There is 1 small blue rubber sphere $\to$ subtract 1. Remaining objects: $10 - 0 - 1 = 9$. \texttt{</think>}
\end{tcolorbox}

\vspace{3pt}

\begin{minipage}[t]{0.48\textwidth}
\begin{tcolorbox}[colback=gray!5, colframe=gray!40, borderline west={2.5pt}{0pt}{red!70}, fontupper=\footnotesize, boxrule=0.3pt, arc=2pt, left=5pt, right=4pt, top=3pt, bottom=3pt]
\textbf{\color{red!70}Generic Cue}\\[2pt]
\texttt{<reground>}
\textcolor{red}{Please look at the image again carefully.}
\texttt{</reground>}

\texttt{<think>} Re-counting all objects from left to right: Small cyan matte cube (far left). Large cyan metallic sphere. Large blue matte cylinder. Small brown matte cube (on top area). Small red matte cube. Large gray matte sphere. Red metallic sphere. Large gold metallic cylinder. Small blue matte sphere. Small cyan matte cylinder (far right). Wait --- now I need to reconsider. The small cyan cylinder on the far right... could it be considered a ``tiny yellow matte cylinder''? Looking again, its color appears to be a very light cyan, but under the lighting conditions it could have a slight yellowish tint. The object is small enough to qualify as ``tiny'' and it does appear to be matte. Actually, I think I may have been too hasty. \textcolor{red}{Let me reconsider --- the cylinder on the right has a pale, somewhat warm-toned surface. It could plausibly be interpreted as a light yellow or cream color rather than cyan.} If so, it would count as a tiny yellow matte cylinder. Also, I want to double-check: is the small blue sphere truly ``rubber''? The surface looks quite smooth and could potentially be shiny rather than matte. But the question says ``rubber'' which typically implies matte, so I will keep counting it.
Total objects: 10.
- Subtract tiny yellow matte cylinders: \textcolor{red}{1} (the small cylinder on the right, which I now believe may be yellow rather than cyan)
- Subtract small blue rubber spheres: 1.
Remaining: $10 - 1 - 1 = 8$ \texttt{</think>}

\textbf{Answer: 8} \ding{55} \textcolor{red}{(negative flip)}
\end{tcolorbox}
\end{minipage}
\hfill
\begin{minipage}[t]{0.48\textwidth}
\begin{tcolorbox}[colback=gray!5, colframe=gray!40, borderline west={2.5pt}{0pt}{teal!80}, fontupper=\footnotesize, boxrule=0.3pt, arc=2pt, left=5pt, right=4pt, top=3pt, bottom=3pt]
\textbf{\color{teal!80}Targeted Self-Diagnosis}\\[2pt]
\texttt{<reground>}
\textcolor{teal}{Wait, I need to re-examine the small blue sphere --- is it truly matte and blue, or could it be shiny like the other spheres? Also, let me check if there's any yellow object that might have been misidentified as a cylinder.}
\texttt{</reground>}

\texttt{<think>} The small blue sphere is indeed matte and blue --- its surface reflects very little light, confirming it is non-shiny. It is located near the center-right of the image, slightly behind the large gold cylinder. \textcolor{teal}{There are no tiny yellow matte cylinders in the image.}
Total objects = 10 (as previously noted). Identify objects to subtract.
Subtract all tiny yellow matte cylinders: After re-examination, no such object exists $\to$ subtract 0.
Subtract all small blue rubber spheres: Only one small blue matte sphere is present $\to$ subtract 1. Apply subtraction.
Remaining objects = Total $-$ (yellow cylinders) $-$ (blue spheres) = $10 - 0 - 1 = 9$. Rule applied: Simple subtraction based on presence/absence of specified objects. \texttt{</think>}

\textbf{Answer: 9} \ding{51} \textcolor{teal}{(confirmed)}
\end{tcolorbox}
\end{minipage}

\caption{Generic vs.\ targeted diagnostic cue on a CLEVR-Math spatial reasoning problem.
Both conditions share the same correct Round~1 (answer: 9).
\textbf{Left:} The generic cue triggers unfocused re-counting; the model hallucinated a yellow cylinder by reinterpreting the cyan cylinder's color under lighting conditions, overriding the correct answer (negative flip: $9\to 8$~\ding{55}).
\textbf{Right:} The targeted cue directs verification to specific object attributes (material, color), confirming the correct count ($9$~\ding{51}).}
\label{fig:case_generic}
\end{figure*}

Figure~\ref{fig:case_generic} compares generic and targeted re-examination on a CLEVR-Math spatial reasoning problem.
Both conditions share the same correct Round~1 (answer: 9).
The generic cue (``Please look at the image again carefully'') triggers unfocused re-counting: the model second-guesses the color of the cyan cylinder, hallucinating it as yellow, and overrides its correct answer (negative flip: $9 \to 8$).
In contrast, the targeted cue directs attention to two specific attributes---the blue sphere's material and the presence of yellow objects---leading to focused verification that confirms the original count.
This example directly illustrates the cue-quality ablation in Table~5 of the main text: generic cues drop MathVista by $-0.8$ and HallusionBench by $-2.1$ below the no-re-examination baseline, while targeted cues yield gains of $+3.8$ and $+6.8$ respectively.

\subsection{Cross-Domain Examples}

We present three additional trajectory types: verification (Figure~\ref{fig:case_geometry}), correction (Figure~\ref{fig:case_correction}), and No-ReGround (Figure~\ref{fig:case_chartmed}).

Figure~\ref{fig:case_geometry} shows a verification trajectory on a MathVista problem (\#744) where Round~1 correctly applies the Basic Proportionality Theorem.
The diagnostic cue targets the parallelism assumption and the proportion setup.
After image re-injection, Round~2 confirms both the arrow-based parallelism and the algebraic derivation, retaining the correct answer.
This demonstrates that re-examination can serve as confirmation rather than correction---an important behavior taught by the Step~3 (Verification) trajectories in our training curriculum.

Figure~\ref{fig:case_correction} shows a correction trajectory on an MMStar problem (\#671) where Round~1 misidentifies a sub-area as the full anatomical structure.
The diagnostic cue directs the model to re-examine the label hierarchy in the brain diagram. After image re-injection, Round~2 correctly distinguishes the motor cortex (a sub-area) from the frontal lobe (the full lobe), flipping from incorrect to correct.

Figure~\ref{fig:case_chartmed} presents two No-ReGround examples where the model produces correct, visually grounded answers in a single round without triggering re-examination, demonstrating that the trained policy learns appropriate restraint.

\begin{figure*}[t]
\centering

\begin{tcolorbox}[colback=gray!8, colframe=gray!40, boxrule=0.4pt, arc=2pt, left=5pt, right=5pt, top=4pt, bottom=4pt]
\begin{minipage}[c]{0.58\textwidth}
\small
\textbf{Q:} If RL $= 5$, RT $= 9$, and WS $= 6$, find RW.\\
(A) 5 \quad (B) 5.4 \quad (C) 6.6 \quad (D) 7.5\\[3pt]
\colorbox{gray!15}{\footnotesize Source: MathVista \#744}
\end{minipage}
\hfill
\begin{minipage}[c]{0.34\textwidth}
\centering
\includegraphics[width=0.95\textwidth]{figures/case_geometry_image.png}
\end{minipage}
\end{tcolorbox}

\vspace{3pt}

\begin{tcolorbox}[colback=gray!5, colframe=gray!40, borderline west={2.5pt}{0pt}{teal!70}, boxrule=0.3pt, fontupper=\footnotesize, left=5pt, right=5pt, top=4pt, bottom=4pt]
\textbf{Round~1} \texttt{<think>}
The diagram shows a triangle RST with points L on RT and W on RS. Segments LW and TS are marked with pink arrows, indicating they are parallel (LW $\parallel$ TS). Points are labeled R, L, T, W, S. Given values: RL $= 5$, RT $= 9$, WS $= 6$.
The diagram suggests triangle RST with line segment LW parallel to side TS, intersecting sides RT and RS. Since LW $\parallel$ TS, by the Basic Proportionality Theorem (also known as Thales' Theorem or the Triangle Proportionality Theorem), we have: if a line is drawn parallel to one side of a triangle intersecting the other two sides, then it divides those sides proportionally. In triangle RST, with LW $\parallel$ TS, we get: $\text{RL}/\text{LT} = \text{RW}/\text{WS}$.
We are given: RL $= 5$, RT $= 9$, so LT $= \text{RT} - \text{RL} = 9 - 5 = 4$, and WS $= 6$. Let RW $= x$. Substitute into the proportion: $5/4 = x/6$. Cross-multiply: $5 \times 6 = 4 \times x$, so $30 = 4x$, $x = 30/4$, $x = 7.5$. Therefore, RW $= 7.5$. This matches option D.
\texttt{</think>} \textbf{Answer: D} \ding{51}

\texttt{<reground>} \textcolor{teal}{Wait, I need to re-check if segments LW and TS are truly parallel based on the direction of the pink arrows --- they seem to indicate parallelism, but I should confirm if that's consistent with the figure's orientation. Also, let me verify if the proportion between RL and RT applies correctly to RW and RS under that parallel assumption.} \texttt{</reground>}

\textbf{Round~2} \texttt{<think>}
My initial assumption that LW $\parallel$ TS comes from the pink arrows, and upon re-checking, both segments share the same arrow style and direction, which is consistent with parallel lines. The diagram shows triangle RST with L on RT and W on RS, and segment LW connecting them.
Since LW $\parallel$ TS, we can apply the Basic Proportionality Theorem: a line parallel to one side of a triangle divides the other two sides proportionally. Given RL $= 5$ and RT $= 9$, we get LT $= 9 - 5 = 4$. Also WS $= 6$. Let RW $= x$. Using the proportion $\text{RL}/\text{LT} = \text{RW}/\text{WS}$: $5/4 = x/6 \Rightarrow x = 7.5$, matching option D.
\texttt{</think>} \textbf{Answer: D} \ding{51} \textcolor{teal}{(verified)}
\end{tcolorbox}

\caption{Verification example (MathVista \#744). Round~1 correctly applies the Basic Proportionality Theorem. The diagnostic cue targets the parallelism assumption (LW $\parallel$ TS) and the proportion setup. After image re-injection, Round~2 confirms both and retains the correct answer (D: 7.5).}
\label{fig:case_geometry}
\end{figure*}

\begin{figure*}[t]
\centering
\begin{tcolorbox}[colback=gray!8, colframe=gray!40, boxrule=0.4pt, arc=2pt, left=5pt, right=5pt, top=4pt, bottom=4pt]
\begin{minipage}[c]{0.62\textwidth}
\small
\textbf{Q:} Which part of the human brain is the largest and most anterior part of each cerebral hemisphere?\\
(A) motor cortex \quad (B) occipital lobe \quad (C) temporal lobe \quad (D) frontal lobe\\[2pt]
\colorbox{gray!15}{\footnotesize Source: MMStar \#671}
\end{minipage}
\hfill
\begin{minipage}[c]{0.30\textwidth}
\centering
\includegraphics[width=0.95\textwidth]{figures/case_brain_image.png}
\end{minipage}
\end{tcolorbox}

\vspace{3pt}

\begin{tcolorbox}[colback=gray!5, colframe=gray!40, borderline west={2.5pt}{0pt}{red!70}, boxrule=0.3pt, fontupper=\footnotesize, left=5pt, right=5pt, top=4pt, bottom=4pt]
\textbf{Round~1} \texttt{<think>}
The image shows a lateral view of the human brain with four lobes labeled: Frontal lobe (purple), Parietal lobe (beige), Temporal lobe (pink), and Occipital lobe (blue). Below the brain, text explains that deep fissures divide the cortex into four regions.
Looking at the diagram, the motor cortex is prominently labeled in the purple region at the front of the brain. It appears to occupy the anterior portion and is visually the most prominent labeled area in the frontal region. The question asks for the ``largest and most anterior part,'' and the motor cortex label is clearly positioned at the front.
(B)~Occipital lobe is at the back --- not anterior.
(C)~Temporal lobe is below and lateral --- not the most anterior.
(D)~Frontal lobe overlaps with motor cortex, but motor cortex is the specific functional area labeled at the front.
\textcolor{red}{The motor cortex is the most anterior labeled structure, so I will choose (A).}
\texttt{</think>} \textbf{Answer: A} \ding{55}

\texttt{<reground>} \textcolor{teal}{Wait, I need to re-examine whether ``motor cortex'' refers to the entire front region of the brain or just a narrow functional strip within a larger lobe. Let me look more carefully at how the labels are organized --- is there a broader lobe label that encompasses the motor cortex area?} \texttt{</reground>}

\textbf{Round~2} \texttt{<think>}
Re-examining the diagram more carefully, I can now see that the large purple region is labeled ``Frontal lobe'' as a whole, and ``Motor cortex'' is a specific sub-area label pointing to a narrow strip at the posterior boundary of the frontal lobe. The frontal lobe extends from the forehead all the way back to the central fissure, making it the largest lobe in terms of surface area. The motor cortex, by contrast, is just one functional zone within it.
The question asks for the ``largest and most anterior \emph{part}'' --- this refers to the frontal lobe as an anatomical division, not a specific functional sub-area.
(A)~Motor cortex is a functional area \emph{within} the frontal lobe, not the entire lobe.
(D)~Frontal lobe is the correct answer --- it is both the largest lobe and the most anteriorly positioned.
\texttt{</think>} \textbf{Answer: D} \ding{51} \textcolor{teal}{(corrected)}
\end{tcolorbox}

\caption{Correction example (MMStar \#671). Round~1 confuses the motor cortex (a functional sub-area) with the frontal lobe (the full anatomical division), selecting the wrong answer (A). The diagnostic cue directs re-examination of the label hierarchy---whether ``motor cortex'' is the entire front region or a sub-area within a larger lobe. After image re-injection, Round~2 correctly identifies that the frontal lobe encompasses the motor cortex and is the largest, most anterior part, flipping to the correct answer (D).}
\label{fig:case_correction}
\end{figure*}

\begin{figure*}[t]
\centering

\noindent\textbf{\footnotesize (a) MathVista \#314 --- Bar chart reading}
\vspace{2pt}

\begin{tcolorbox}[colback=gray!8, colframe=gray!40, boxrule=0.4pt, arc=2pt, left=5pt, right=5pt, top=4pt, bottom=4pt]
\begin{minipage}[c]{0.55\textwidth}
\small
\textbf{Q:} Is Crimson the minimum? \quad Choices: (A) yes \quad (B) no\\[2pt]
\colorbox{gray!15}{\footnotesize Source: MathVista \#314}
\end{minipage}
\hfill
\begin{minipage}[c]{0.38\textwidth}
\centering
\includegraphics[width=0.85\textwidth]{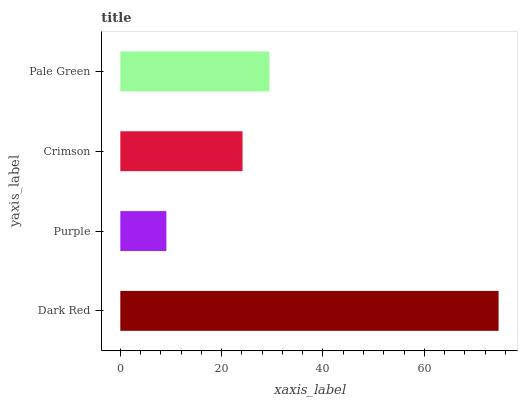}
\end{minipage}
\end{tcolorbox}
\vspace{2pt}
\begin{tcolorbox}[colback=gray!5, colframe=gray!30, boxrule=0.3pt, fontupper=\footnotesize, left=5pt, right=5pt, top=3pt, bottom=3pt]
\texttt{<think>} The image is a horizontal bar chart. The x-axis is labeled ``xaxis\_label'' and ranges from 0 to 70, with tick marks at intervals of 10 (0, 10, 20, 30, 40, 50, 60, 70). The y-axis is labeled ``yaxis\_label'' and has four categories listed vertically: ``Dark Red'', ``Purple'', ``Crimson'', ``Pale Green''. Each category has a horizontal bar extending to the right from the y-axis. The bar for ``Dark Red'' extends to approximately 75 on the x-axis. The bar for ``Pale Green'' extends to approximately 30. The bar for ``Crimson'' extends to approximately 25. The bar for ``Purple'' extends to approximately 15. The title of the chart is ``title''.
To determine if ``Crimson'' is the minimum, we compare the lengths of all bars. The bar for ``Dark Red'' is the longest, extending beyond 70. The bar for ``Pale Green'' is shorter than ``Dark Red'' but longer than ``Crimson''. The bar for ``Crimson'' is shorter than ``Pale Green''. The bar for ``Purple'' is the shortest, extending to approximately 15. Therefore, ``Purple'' is the minimum, not ``Crimson''. So, the answer to ``Is Crimson the minimum?'' is ``no''. \texttt{</think>}\\
\textbf{Answer: B} \ding{51} \quad \textit{No \texttt{<reground>} triggered --- single-round, visually grounded.}
\end{tcolorbox}

\vspace{8pt}

\noindent\textbf{\footnotesize (b) MathVista \#563 --- Medical monitor}
\vspace{2pt}

\begin{tcolorbox}[colback=gray!8, colframe=gray!40, boxrule=0.4pt, arc=2pt, left=5pt, right=5pt, top=4pt, bottom=4pt]
\begin{minipage}[c]{0.55\textwidth}
\small
\textbf{Q:} Which number on the monitor is higher?\\
(A) top \quad (B) bottom \quad (C) left \quad (D) right\\[2pt]
\colorbox{gray!15}{\footnotesize Source: MathVista \#563}
\end{minipage}
\hfill
\begin{minipage}[c]{0.38\textwidth}
\centering
\includegraphics[width=0.85\textwidth]{figures/case_medical_image.png}
\end{minipage}
\end{tcolorbox}
\vspace{2pt}
\begin{tcolorbox}[colback=gray!5, colframe=gray!30, boxrule=0.3pt, fontupper=\footnotesize, left=5pt, right=5pt, top=3pt, bottom=3pt]
\texttt{<think>} The monitor in the image displays two prominent numbers: one at the top of the screen and one at the bottom. The top number is ``73'' and is located above the ECG waveform. The bottom number is ``97'' and is positioned below the ECG waveform, aligned with the heart rate display. Both numbers are clearly visible on the monitor's screen. The question asks which number is higher.
Option (A)~top: The top number is 73.
Option (B)~bottom: The bottom number is 97.
Comparing the two: $97 > 73$.
Therefore, the bottom number is higher.
Options (C)~left and (D)~right refer to positions on the screen, not specific numbers. The left side shows other parameters like blood pressure, but the question specifically asks about the two main numbers displayed.
Since 97 is greater than 73, the correct answer is (B)~bottom. \texttt{</think>}\\
\textbf{Answer: B} \ding{51} \quad \textit{No \texttt{<reground>} triggered --- single-round, visually grounded.}
\end{tcolorbox}

\caption{No-ReGround examples. \textbf{Top:}~MathVista \#314 --- bar chart reading: the model reads all four bar values from the chart axes, compares them systematically, and correctly determines that Purple ($\sim$15) is the minimum rather than Crimson ($\sim$25). \textbf{Bottom:}~MathVista \#563 --- medical monitor: the model identifies both numerical values (73 and 97), eliminates the irrelevant spatial options (C and D), and correctly concludes that the bottom number is larger. Both examples are solved in a single round without triggering re-examination, demonstrating that the policy learns appropriate restraint.}
\label{fig:case_chartmed}
\end{figure*}

\subsection{Summary of Qualitative Patterns}

Across all qualitative examples above and in Figures~3--4 of the main text, we consistently observe two patterns.

\noindent\textbf{Pattern 1: Targeted cues produce focused correction.}
Diagnostic cues that reference specific visual elements (object attributes, geometric marks, chart labels) produce focused Round~2 reasoning that either corrects errors or confirms correct answers. In the correction example (Figure~\ref{fig:case_correction}), the cue specifically asks whether ``motor cortex'' is a sub-area or the full region, leading the model to re-examine the label hierarchy and flip to the correct answer. In the verification example (Figure~\ref{fig:case_geometry}), the cue targets the parallelism assumption, and the model confirms it through careful re-reading of the arrow markers. The attention redistribution is semantically aligned with the cue content (cf.\ the SAS analysis above).

\noindent\textbf{Pattern 2: Generic cues produce harmful second-guessing.}
Undirected re-examination triggers diffuse re-counting or re-interpretation that frequently overrides correct first-round answers. In the CLEVR-Math example (Figure~\ref{fig:case_generic}), the generic cue leads the model to hallucinate a non-existent yellow cylinder by reinterpreting the cyan cylinder's color under lighting conditions, causing a negative flip. This pattern is consistent with our finding that generic cues actively degrade accuracy on MathVista ($-0.8$) and HallusionBench ($-2.1$) relative to the no-re-examination baseline.

These two patterns are consistent with the quantitative cue-quality ablation in Table~5 of the main paper and the flip-rate decomposition in Supplementary Table~\ref{tab:fliprate}, where the best positive-to-negative flip ratio (3.4$\times$) occurs at the optimal trigger rate of 45\%.

\end{document}